\documentclass[sigconf]{acmart}
\AtBeginDocument{%
  \providecommand\BibTeX{{%
    \normalfont B\kern-0.5em{\scshape i\kern-0.25em b}\kern-0.8em\TeX}}}

\copyrightyear{2024}
\acmYear{2024}
\setcopyright{acmlicensed}
\acmConference[CIKM '24] {Proceedings of the 33rd ACM International Conference on Information and Knowledge Management}{October 21--25, 2024}{Boise, ID, USA.}
\acmBooktitle{Proceedings of the 33rd ACM International Conference on Information and Knowledge Management (CIKM '24), October 21--25, 2024, Boise, ID, USA}
\acmISBN{979-8-4007-0436-9/24/10}
\acmDOI{10.1145/3627673.3679847}

\usepackage[utf8]{inputenc} 
\usepackage[T1]{fontenc}    
\usepackage{hyperref}       
\usepackage{url}            
\usepackage{booktabs}       
\usepackage{amsfonts}       
\usepackage{nicefrac}       
\usepackage{microtype}      
\usepackage{xcolor}         
\usepackage{graphicx}
\usepackage{subfigure}
\usepackage{cleveref}
\usepackage{caption}
\usepackage{threeparttable}
\usepackage{comment}
\usepackage{multirow}
\usepackage{enumitem}
\usepackage{adjustbox}
\usepackage{bbding}
\usepackage[normalem]{ulem}
\useunder{\uline}{\ul}{}
\usepackage{latexsym}
\usepackage{algorithm}
\usepackage{algorithmic}
\usepackage{mathrsfs}
\usepackage{balance}
\usepackage{dcolumn}
\usepackage{tabularx}
\usepackage{colortbl}
\usepackage{xspace,mfirstuc,tabulary}
\usepackage{amsthm}
\usepackage{ulem}
\usepackage{multirow,booktabs, hhline}
\usepackage{bm}
\usepackage{color}
\newenvironment{itemize*}%
 {\leftmargini=20pt\begin{itemize}%
  \setlength{\itemsep}{3pt}%
  \setlength{\parskip}{0pt}%
  }%
 {\end{itemize}}
\newenvironment{enumerate*}%
 {\begin{enumerate}%
  \setlength{\itemsep}{0pt}%
  \setlength{\parskip}{0pt}}%
 {\end{enumerate}}
\usepackage{amsmath}
\usepackage{makecell}
\usepackage{bbm}
\usepackage{CJKutf8}
\usepackage{listings}
\usepackage{wrapfig}
\usepackage{lipsum}
\usepackage{pgfplots}
\usepackage{tikz}
\usetikzlibrary{calc}
\usepackage{array}
\usepackage{arydshln}
\usepackage{multicol}

\definecolor{chart Idle}{gray}{.6}
\definecolor{chart Poor}{RGB}{242,28,28}
\definecolor{chart Ok}{RGB}{248,172,37}
\definecolor{chart Ideal}{RGB}{1,151,0}
\definecolor{chart Over}{RGB}{0,125,234}
\definecolor{lightergray}{RGB}{230,230,230}
\definecolor{DarkGreen}{RGB}{30,130,30}

\newcommand\ourdata{ToolLens\xspace}
\newcommand\ourmodel{COLT\xspace}
\newcommand{\paratitle}[1]{\vspace{1.5ex}\noindent\textbf{#1}}
\newcommand{\ie}{\emph{i.e.,}\xspace}

\newcommand{\wo}{\emph{w/o}\xspace}

\newcommand\oureval{COMP}
\newcommand\Oureval{COMP\xspace}
\newdimen\tempdim

\newcommand*{\ChartBox}[3]{%
  \begingroup
    \settoheight{\tempdim}{L}%
    \edef\tempheight{\the\tempdim}%
    \settodepth{\tempdim}{g}%
    \edef\tempdepth{\the\tempdim}%
    \tikz[
      baseline=0pt,
      inner sep=0pt,
    ]
    \node[
      fill={#3!#2},
      rounded corners=1pt,
      anchor=base,
    ]{%
      \vphantom{g\"A}%
      \pgfmathsetlength{\tempdim}{#1}%
      \kern\tempdim\relax
    };%
  \endgroup
}
\newcommand*{\chart}[3]{%
  \ChartBox{20mm/3000*(#1-400)}{#2}{#3}%
}

\AtBeginDocument{%
  \providecommand\BibTeX{{%
    \normalfont B\kern-0.5em{\scshape i\kern-0.25em b}\kern-0.8em\TeX}}}
\begin{document}
\begin{sloppy}
\title{Towards Completeness-Oriented Tool Retrieval for\\ Large Language Models}

\author{Changle Qu}
\author{Sunhao Dai}
\affiliation{
  \institution{\mbox{Gaoling School of Artificial Intelligence}\\Renmin University of China}
  \city{Beijing}
  \country{China}
  }
\email{{changlequ,sunhaodai}@ruc.edu.cn}

\author{Xiaochi Wei}
\affiliation{
  \institution{Baidu Inc.}
  \city{Beijing}
  \country{China}
  }
\email{weixiaochi@baidu.com}

\author{Hengyi Cai}
\affiliation{%
  \institution{Institute of Computing Technology\\ Chinese Academy of Sciences}
  \city{Beijing}
  \country{China}
  }
\email{caihengyi@ict.ac.cn}

\author{Shuaiqiang Wang}
\affiliation{
  \institution{Baidu Inc.}
  \city{Beijing}
  \country{China}
  }
\email{shqiang.wang@gmail.com}

\author{Dawei Yin}
\affiliation{
  \institution{Baidu Inc.}
  \city{Beijing}
  \country{China}
  }
\email{yindawei@acm.org}

\author{Jun Xu}
\authornote{Jun Xu is the corresponding author. Work partially done at Engineering Research Center of Next-Generation Intelligent Search and Recommendation, Ministry of Education.}
\author{Ji-Rong Wen}
\affiliation{
  \institution{\mbox{Gaoling School of Artificial Intelligence}\\Renmin University of China}
  \city{Beijing}
  \country{China}
  }
\email{{junxu,jrwen}@ruc.edu.cn}

\renewcommand{\authors}{Changle Qu, Sunhao Dai, Xiaochi Wei, Hengyi Cai, Shuaiqiang Wang, Dawei Yin, Jun Xu, and Ji-Rong Wen}
\renewcommand{\shorttitle}{Towards Completeness-Oriented Tool Retrieval for Large Language Models}
\renewcommand{\shortauthors}{Changle Qu et al.}
\begin{abstract}
Recently, integrating external tools with Large Language Models (LLMs) has gained significant attention as an effective strategy to mitigate the limitations inherent in their pre-training data. However, real-world systems often incorporate a wide array of tools, making it impractical to input all tools into LLMs due to length limitations and latency constraints. Therefore, to fully exploit the potential of tool-augmented LLMs, it is crucial to develop an effective tool retrieval system. Existing tool retrieval methods primarily focus on semantic matching between user queries and tool descriptions, frequently leading to the retrieval of redundant, similar tools. Consequently, these methods fail to provide a complete set of diverse tools necessary for addressing the multifaceted problems encountered by LLMs. In this paper, we propose a novel model-agnostic \textbf{\underline{C}}\textbf{\underline{O}}llaborative \textbf{\underline{L}}earning-based \textbf{\underline{T}}ool Retrieval approach, \textbf{\ourmodel}, which captures not only the semantic similarities between user queries and tool descriptions but also takes into account the collaborative information of tools. Specifically, we first fine-tune the PLM-based retrieval models to capture the semantic relationships between queries and tools in the semantic learning stage. Subsequently, we construct three bipartite graphs among queries, scenes, and tools and introduce a dual-view graph collaborative learning framework to capture the intricate collaborative relationships among tools during the collaborative learning stage. Extensive experiments on both the open benchmark and the newly introduced \ourdata dataset show that \ourmodel achieves superior performance. Notably, the performance of BERT-mini (11M) with our proposed model framework outperforms BERT-large (340M), which has 30 times more parameters. Furthermore, we will release \ourdata publicly to facilitate future research on tool retrieval.
\end{abstract}



\maketitle
 
\section{Introduction}
\label{introsection}
Recently, large language models~(LLMs) have demonstrated remarkable progress across various natural language processing tasks~\cite{brown2020language,chowdhery2023palm,achiam2023gpt,touvron2023llama}.
However, they often struggle with solving highly complex problems and providing up-to-date knowledge due to the constraints of their pre-training data~\cite{mallen2022not,vu2023freshllms}.
A promising approach to overcome these limitations is tool learning~\cite{schick2023toolformer,parisi2022talm,li-etal-2023-api,ye2024tooleyes,qu2024tool}, which enables LLMs to dynamically interact with external tools, significantly facilitating access to real-time data and the execution of complex computations. 
By integrating tool learning, LLMs transcend the confines of their outdated or limited pre-trained knowledge~\cite{brown2020language}, offering responses to user queries with significantly improved accuracy and relevance~\cite{huang2023metatool, qin2023toolllm}.
However, real-world systems typically involve a large number of tools, making it impractical to take the descriptions of all tools as input for LLMs due to length limitations and latency constraints. Thus, as illustrated in Figure~\ref{fig:intro_process}, developing an effective tool retrieval system becomes essential to fully exploit the potential of tool-augmented LLMs~\cite{Gao2023ConfuciusIT}.

\begin{figure}[t]
\centering
\subfigure[Pipeline of user interaction with tool-augmented LLMs.]
{
    \includegraphics[width=0.85\linewidth]{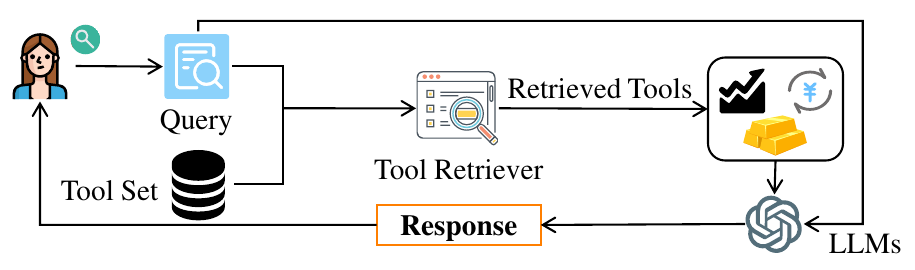}
    \label{fig:intro_process}
}
\subfigure[Illustration of different responses with different tools.]
{
    \includegraphics[width=0.85\linewidth]{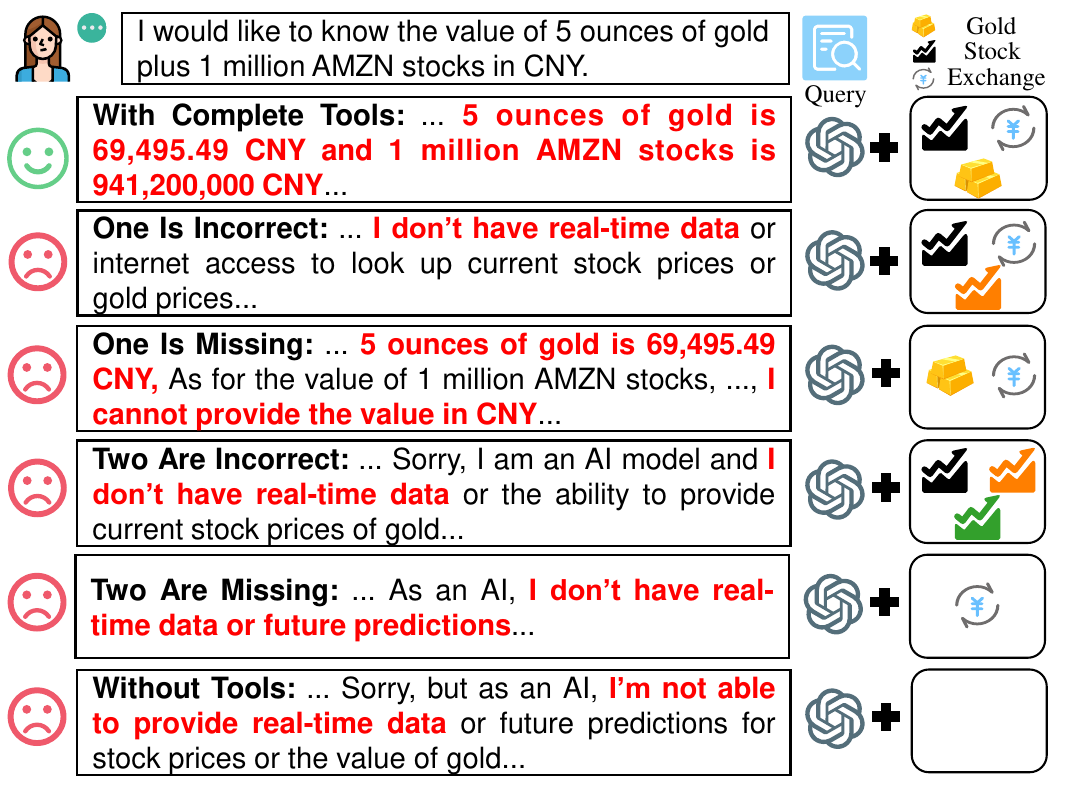}
    \label{fig:intro_case}
}
\caption{An illustration of tool retrieval for LLMs with tool learning and varied responses using different tools.}
\vspace{-2em}
\label{fig:intro}
\end{figure}

Typically, existing tool retrieval approaches directly employ dense retrieval techniques~\cite{qin2023toolllm,Gao2023ConfuciusIT,yuan2024easytool}, solely focusing on matching semantic similarities between queries and tool descriptions.
However, these approaches may fall short when addressing multifaceted queries that require a collaborative effort from multiple tools to formulate a comprehensive response.
For instance, in Figure~\ref{fig:intro_case}, consider a user's request to calculate the value of 5 ounces of gold plus 1 million AMZN stocks in CNY. 
Such a query requires the simultaneous use of tools for gold prices, stock values, and currency exchange rates. 
The absence of any of these tools yields an incomplete answer. In this example, dense retrieval methods that rely solely on semantic matching may retrieve multiple tools related to stock prices while neglecting others. This highlights a significant limitation of dense retrieval methods that overlook the necessity for tools to interact collaboratively.
Thus, ensuring the completeness of retrieved tools is an essential aspect of a tool retrieval system, which is often neglected by traditional retrieval approaches.

Toward this end, this paper proposes \textbf{\ourmodel}, a novel model-agnostic \textbf{\underline{C}}\textbf{\underline{O}}llaborative \textbf{\underline{L}}earning-based \textbf{\underline{T}}ool retrieval approach aimed at enhancing completeness-oriented tool retrieval.
This method is structured into two main stages: semantic learning and collaborative learning.
Initially, we fine-tune traditional pre-trained language models (PLMs) on tool retrieval datasets to acquire semantic matching information between queries and tools, thereby addressing the potential performance issues of these models in zero-shot scenarios for tool retrieval tasks.
Subsequently, to capture the intricate collaborative relationship among tools, a concept of ``scene'' is proposed to indicate a group of collaborative tools. 
Based on this, \ourmodel integrates three bipartite graphs among queries, scenes, and tools. 
More specifically, given the initial semantic embedding from the semantic learning stage, the high-order collaborative relationship is better integrated via message propagation and cross-view graph contrastive learning among these graphs.
The learning objective incorporates a list-wise multi-label loss to ensure the simultaneous acquisition of tools from the entire ground-truth set without favoring any specific tool.

Moreover, traditional retrieval metrics like Recall~\cite{zhu2004recall} and NDCG~\cite{jarvelin2002cumulated} fail to capture the completeness necessary for effective tool retrieval. 
As illustrated in Figure~\ref{fig:intro_case}, the exclusion of any essential tool from the ground-truth tool set compromises the ability to fully address user queries, indicating that metrics focused solely on individual tool ranking performance are inadequate when multiple tools are required. 
To bridge this gap, we introduce \oureval@$K$, a new metric designed to assess tool retrieval performance based on completeness, which can serve as a reliable indicator of how well a tool retrieval system performs for downstream tool learning applications. 
Additionally, we construct a new dataset called \ourdata, in which a query is typically solved with multiple relevant but diverse tools, reflecting the multifaceted nature of user requests in real-world scenarios. 

In summary, our main contributions are as follows: 

\textbullet~The collaborative relationships among multiple tools in LLMs have been thoroughly studied, which reveals that incomplete tool retrieval hinders accurate answers, underscoring the integral role each tool plays in the collective functionality. 

\textbullet~We introduce \ourmodel, a novel tool retrieval approach that uses message propagation and cross-view graph contrastive learning among queries, scenes, and tools, incorporating better collaborative information among various tools.

\textbullet~Extensive experiments demonstrate the superior performance of \ourmodel against state-of-the-art dense retrieval methods in both tool retrieval and downstream tool learning. 

\textbullet~We introduce a new dataset and a novel evaluation metric specifically designed for assessing multi-tool usage in LLMs, which will facilitate future research on tool retrieval.

\section{Related Work}
\paratitle{Tool Learning.} 
Recent studies highlight the potential of LLMs to utilize tools in addressing complex problems~\citep{qin2023tool,mialon2023augmented}. 
Existing tool learning approaches can be categorized into two types: tuning-free and tuning-based methods~\citep{Gao2023ConfuciusIT}.
Tuning-free methods capitalize on the in context learning ability of LLMs through strategic prompting~\cite{wei2022chain,yao2022react,song2023restgpt,shen2024hugginggpt}.
For example, ART~\cite{paranjape2023art} constructs a task library, from which it retrieves demonstration examples as few-shot prompts when encountering real-world tasks.
Conversely, tuning-based methods involve directly fine-tuning the parameters of LLMs on specific tool datasets to master tool usage.
For example, ToolLLaMA~\cite{qin2023toolllm} employs the instruction-solution pairs derived from the DFSDT method to fine-tune the LLaMA model, thereby significantly enhancing its tool usage capabilities.
Despite these advancements, most strategies either provide a manual tool set~\citep{schick2023toolformer,tang2023toolalpaca,xu2023tool} or employ simple dense retrieval~\citep{Gao2023ConfuciusIT} for tool retrieval. 
However, LLMs must choose several useful tools from a vast array of tools in real-world applications, necessitating a robust tool retriever to address the length limitations and latency constraints of LLMs.

\paratitle{Tool Retrieval.} Tool retrieval aims to find top-$K$ most suitable tools for a given query from a vast set of tools.
Existing tool retrieval methods typically directly adopt traditional retrieval approaches, and state-of-the-art retrieval methods can be categorized into two types: term-based and semantic-based. 
Term-based methods, such as TF-IDF~\citep{sparck1972statistical} and BM25~\citep{robertson2009probabilistic}, prioritize term matching via sparse representations.
Conversely, semantic-based methods, such as ANCE~\cite{xiong2020approximate}, TAS-B~\cite{hofstatter2021efficiently}, coCondensor~\cite{gao2021unsupervised}, and Contriever~\cite{izacard2021unsupervised}, utilize neural networks to learn the semantic relationship between queries and tool descriptions and then calculate the semantic similarity using methods such as cosine similarity. 
Despite these advancements, existing methods for tool retrieval overlook the importance of the collaborative relationship among multiple tools, thereby falling short of meeting the completeness criterion for tool retrieval. Our work seeks to mitigate these issues by collaborative learning that leverages graph neural networks and cross-view contrastive learning among graphs.

\section{Our Approach: \ourmodel}

\begin{figure*}[t]
\centering
	\includegraphics[width=\linewidth]{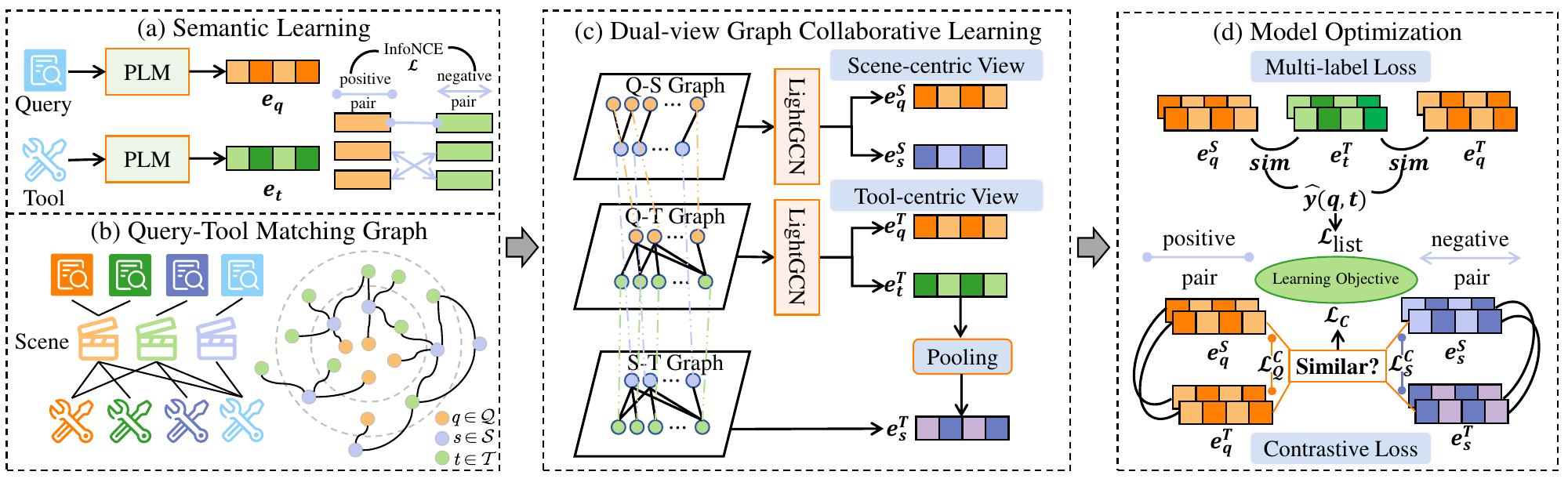}
\caption{The architecture of the proposed two-stage learning framework \ourmodel for tool retrieval.}
        \label{fig:method}
\end{figure*}

In this section, we first introduce task formulation of tool retrieval. Then we describe the details of the proposed \ourmodel approach.

\subsection{Task Formulation}
Formally, given a user query $q \in \mathcal{Q}$, the goal of tool retrieval is to filter out the top-$K$ most suitable tools $\{t^{(1)}, t^{(2)}, \ldots , t^{(K)}\}$ from the entire tool set $\mathcal{T} = \{(t_1,d_1), (t_2,d_2), \ldots, (t_{N},d_{N})\}$, where each element represents a specific tool $t_i$ associated with its description $d_i$, and $N$ is the number of tools in the tool set. 

\paratitle{Goal.}
As discussed in Section \ref{introsection}, the comprehensiveness of the tools retrieved is crucial for LLMs to enhance their ability to accurately address multifaceted and real-time questions. Therefore, it is necessary to ensure that the retrieved tools encompass all the tools required by the user question.
Considering these factors, the goal of tool retrieval is to optimize both accuracy and completeness, ensuring the provision of desired tools for downstream tasks.

\subsection{Overview of \ourmodel}

As illustrated in Figure~\ref{fig:method}, \ourmodel employs a two-stage learning strategy, which includes semantic learning followed by collaborative learning. 
In the first stage, the semantic learning module processes both queries and tools to derive their semantic representations, aiming to align these representations closely within the semantic space.
Subsequently, the collaborative learning module enhances these preliminary representations by introducing three bipartite graphs among queries, scenes, and tools. 
Through dual-view graph contrastive learning within these three bipartite graphs, COLT is able to capture the high-order collaborative information between tools.
Furthermore, a list-wise multi-label loss is utilized in the learning objective to facilitate the balanced retrieval of diverse tools from the complete ground-truth set, avoiding undue emphasis on any specific tool.

In the following sections, we will present the details of these two key learning stages in COLT.

\subsection{Semantic Learning}
\label{semantic learning}

As shown in Figure~\ref{fig:method} (a), in the first stage of COLT, we adopt the established dense retrieval (DR) framework~\cite{zhao2022dense, guo2022semantic}, leveraging pre-trained language models (PLMs) such as BERT~\citep{kenton2019bert} to encode both the query $q$ and the tool $t$ into low dimensional vectors. Specifically, we employ a bi-encoder architecture, with the cosine similarity between the encoded vectors serving as the initial relevance score:
\begin{equation}
\widehat{y}_\text{SL}(q,t)=\mathrm{sim}(\mathbf{e}_q,\mathbf{e}_t), \label{Eq.:encoder}
\end{equation}
where $\mathbf{e}_q$ and $\mathbf{e}_t$ are the mean pooling vectors from the final layer of the PLM, and $\mathrm{sim}(\cdot, \cdot)$ represents the cosine similarity function.

For training, we utilize the InfoNCE loss~\cite{gutmann2010noise,xiong2020approximate}, a standard contrastive learning technique used in training DR models, which contrasts positive pairs against negative ones. 
Specifically, given a query $q$, its relevant tool $t^+$ and the set of irrelevant tools $\{t_{1}^{-}, \cdots, t_{k}^{-}\}$, we minimize the following loss: 
\begin{equation}
-\log \frac{e^{\mathrm{sim}(q, t^+)}}{e^{\mathrm{sim}(q, t^+)} + \sum_{j=1}^{k} e^{\mathrm{sim}(q, t_{j}^{-})}}. \label{Eq.:infonce}
\end{equation}
Through this loss function, we can increase the similarity score between the query and its relevant tool while decreasing the similarity scores between the query and irrelevant tools.

This semantic learning phase ensures good representations for each query and tool from the text description view. However, relying solely on semantic-based retrieval is insufficient for completeness-oriented tool retrieval, as it often falls short in addressing multifaceted queries effectively.

\subsection{Collaborative Learning}
\label{collaborative learning}

\subsubsection{Bipartite Graphs in Tool Retrieval.}
To capture the collaborative information between tools and achieve completeness-oriented tool retrieval, we first formulate the relationship between queries and tools with three bipartite graphs.
Specifically, as illustrated in Figure~\ref{fig:method} (b), we conceptualize the ground-truth tool set for each query as a ``scene'', considering that a collaborative operation of multiple tools is essential to fully address multifaceted queries.
For example, given the query ``I want to travel to Paris.'', it doesn't merely seek a single piece of information but initiates a ``scene'' of travel planning, which involves using various tools for transportation, weather forecasts, accommodation choices, and details about attractions.
This scenario underscores the need for scene matching beyond traditional semantic search or recommendation scenarios, where the focus is on selecting any relevant documents or items without considering their collaborative utility.
As a result, traditional semantic-based retrieval systems may only retrieve tools related to Paris attractions, thus failing to provide a comprehensive and complete tool set for the LLMs. 
Conversely, we construct three bipartite graphs linking queries, scenes, and tools, i.e., Q-S (Query-Scene) graph, Q-T (Query-Tool) graph, and  S-T (Scene-Tool) graph. 
By formulating these three graphs, we can further capture the high-order relationships among tools with graph learning, facilitating a scene-based understanding that aligns to achieve completeness-oriented tool retrieval.

\subsubsection{Dual-view Graph Collaborative Learning.}
Leveraging the initial query and tool representations derived from the first-stage semantic learning, along with the three constructed bipartite graphs, we introduce a dual-view graph collaborative learning framework. This framework is designed to capture the relationships between tools, as depicted in Figure \ref{fig:method} (c). It assesses the relevance between queries and tools from two views: 

\textbf{$\bullet $ Scene-centric View:} Through the Q-S graph and S-T graph, this view captures the relevance between queries and tools mediated by a scene. This offers a nuanced view that considers the collaborative context in which tools work together to meet the requirements of a query.

\textbf{$\bullet $ Tool-centric View:} Utilizing the Q-T graph, this view establishes a direct relevance between each query and its corresponding tools, providing a straightforward measure of their relevance.

This dual-view framework allows for comprehensive access to query-tool relevance, integrating both direct relevance and the broader context of tool collaboration within scenes, thereby enhancing the completeness of tool retrieval.

For the scene-centric view, we adopt the simple but effective Graph Neural Network (GNN)-based LightGCN~\cite{he2020lightgcn} model to delve into the complex relationships between queries and scenes. This is achieved through iterative aggregation of neighboring information across $I$ layers within the Q-S graph. The aggregation process for the $i$-th layer, enhancing the representations of queries $\mathbf{e}_q^{S(i)}$ and scenes $\mathbf{e}_s^{S(i)}$, is defined as follows:
\begin{equation}
\left\{
\begin{aligned}
\mathbf{e}_q^{S(i)} &= \sum_{s \in \mathcal{N}_q^S} \frac{1}{\sqrt{|\mathcal{N}_q^S|}\sqrt{|\mathcal{N}_s^Q|}} \mathbf{e}_s^{S(i-1)}, \\
\mathbf{e}_s^{S(i)} &= \sum_{q \in \mathcal{N}_s^Q} \frac{1}{\sqrt{|\mathcal{N}_q^S|}\sqrt{|\mathcal{N}_s^Q|}} \mathbf{e}_q^{S(i-1)},
\end{aligned}
\right.
\label{Eq.:S_LightGCN}
\end{equation}
where $\mathcal{N}_q^S$, $\mathcal{N}_s^Q$ represent the sets of neighbors of query $q$ and scene $s$ in the Q-S graph, respectively. 
$\mathbf{e}_q^{S(0)}$ originates from the representations acquired in the first semantic learning stage, while $\mathbf{e}_s^{S(0)}$ is derived from the mean pooling of the representations of ground-truth tools associated with each scene:
\begin{equation}
\begin{aligned}
\mathbf{e}_s^{S(0)} = \frac{1}{|\mathcal{N}_s^T|} \sum_{t \in \mathcal{N}_s^T} \mathbf{e}_t,
\end{aligned}
\label{Eq.:e_s_0}
\end{equation}
where $\mathcal{N}_s^T$ represents the set of first-order neighbors of scene $s$ in the S-T graph.

Then we sum the representations from the $0$-th layer to the $I$-th layer to get the final query representations $\mathbf{e}_q^{S}$ and scene representation $\mathbf{e}_s^{S}$ for the scene-centric view:
\begin{equation}
\left\{
\begin{aligned}
\mathbf{e}_q^S &= \mathbf{e}_q^{S(0)} + \dots + \mathbf{e}_q^{S(I)}, \\
\mathbf{e}_s^S &= \mathbf{e}_s^{S(0)} + \dots + \mathbf{e}_s^{S(I)}.
\end{aligned}
\right.
\label{Eq.:S_final}
\end{equation}

In parallel with the scene-centric view, the tool-centric view utilizes LightGCN on the Q-T graph to refine query and tool representations through iterative aggregation.
For each layer $i$, the enhanced representations, $\mathbf{e}_q^{T(i)}$ for queries and $\mathbf{e}_t^{T(i)}$ for tools, are derived as follows:
\begin{equation}
\left\{
\begin{aligned}
\mathbf{e}_q^{T(i)} &= \sum_{t \in \mathcal{N}_q^T} \frac{1}{\sqrt{|\mathcal{N}_q^T|}\sqrt{|\mathcal{N}_t^Q|}} \mathbf{e}_t^{T(i-1)}, \\
\mathbf{e}_t^{T(i)} &= \sum_{q \in \mathcal{N}_t^Q} \frac{1}{\sqrt{|\mathcal{N}_q^T|}\sqrt{|\mathcal{N}_t^Q|}} \mathbf{e}_q^{T(i-1)},
\end{aligned}
\right.
\label{Eq.:T_LightGCN}
\end{equation}
where $\mathcal{N}_q^T$, $\mathcal{N}_t^Q$ represent the first-order neighbors of query $q$ and tool $t$ in the Q-T graph, respectively. $\mathbf{e}_q^{T(0)}$ and $\mathbf{e}_t^{T(0)}$ are obtained from the first semantic learning stage.

Then we sum the representations from the $0$-th layer to the $I$-th layer to derive the final query representations $\mathbf{e}_q^{T}$ and tool representation $\mathbf{e}_t^{T}$ for the tool-centric view:
\begin{equation}
\left\{
\begin{aligned}
\mathbf{e}_q^T &= \mathbf{e}_q^{T(0)} + \dots + \mathbf{e}_q^{T(I)}, \\
\mathbf{e}_t^T &= \mathbf{e}_t^{T(0)} + \dots + \mathbf{e}_t^{T(I)}.
\end{aligned}
\right.
\label{Eq.:T_final}
\end{equation}

Furthermore, leveraging the learned tool representations $\mathbf{e}_t^{T}$ and the S-T graph, the scene representation $\mathbf{e}_s^{T}$ within the tool-centric view can be obtained by pooling all related tool representations:
\begin{equation}
\begin{aligned}
\mathbf{e}_s^{T} = \frac{1}{|\mathcal{N}_s^T|} \sum_{t \in \mathcal{N}_s^T} \mathbf{e}_t^T.
\end{aligned}
\label{Eq.:T_pooling}
\end{equation}

In summary, our dual-view graph collaborative learning framework yields two sets of embeddings: $\mathbf{e}_q^{S}$ and $\mathbf{e}_s^{S}$ from the scene-centric view, and $\mathbf{e}_q^{T}$ and $\mathbf{e}_s^{T}$ from the tool-centric view for queries and scenes, respectively.  Then, the final matching score of each query-tool pair $(q,t)$ is calculated using the following formula:
\begin{equation}
\widehat{y}(q,t)=\mathrm{sim}(\mathbf{e}_q^S,\mathbf{e}_t^T) + \mathrm{sim}(\mathbf{e}_q^T,\mathbf{e}_t^T).
\end{equation}

\begin{figure*}[t]
\centering
	\includegraphics[width=\linewidth]{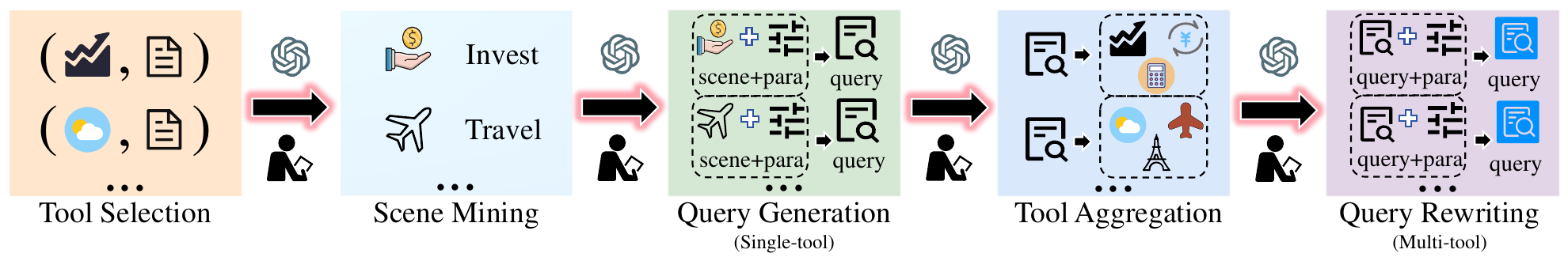}
\caption{An overview of the dataset construction pipeline of \ourdata. Human verification is included at each step.}
        \label{fig:data}
\end{figure*}

\subsubsection{Learning Objective.} 

\begin{algorithm}[t]
\small
\begin{algorithmic}[1]
\caption{The Learning Algorithm of \ourmodel}
\label{alg:COLT}
\renewcommand{\algorithmicrequire}{\textbf{Input:}}
\renewcommand{\algorithmicensure}{\textbf{Output:}}
\newcommand{\PARAMETER}[1]{\item[\textbf{Parameter:}] #1}
\resizebox{1\linewidth}{!}{%
\begin{minipage}{\linewidth}
\REQUIRE PLM, semantic learning training epoch $E$, Query-scene bipartite graph, query-tool bipartite graph, scene-tool bipartite graph, learning rate $lr$, weight decay, layer number $I$, contrastive loss weight $\lambda$, temperature coefficient $\tau$, list length $L$.
\ENSURE \ourmodel Model with learnable parameters $\theta$.\\
\texttt{// Semantic Learning:}
\FOR{$e=1$ \TO $E$}
    \STATE Calculate the InfoNCE loss using Eq.~(\ref{Eq.:infonce})
    \STATE Update parameter of PLM using AdaW
\ENDFOR\\
\texttt{// Collaborative Learning:}
\STATE Calculate initial $\mathbf{e}_q^{S(0)}$, $\mathbf{e}_s^{S(0)}$, $\mathbf{e}_q^{T(0)}$ and $\mathbf{e}_t^{T(0)}$ using the embeddings obtained from the first-stage semantic learning and Eq.~(\ref{Eq.:e_s_0})
\WHILE{COLT not Convergence}
    \FOR{$i=1$ \TO $I$}
    \STATE Conduct message propagation using Eq.~(\ref{Eq.:S_LightGCN}) and Eq.~(\ref{Eq.:T_LightGCN})
    \ENDFOR
    \STATE Calculate final $\mathbf{e}_q^{S}$, $\mathbf{e}_s^{S}$, $\mathbf{e}_q^{T}$, $\mathbf{e}_s^{T}$ and $\mathbf{e}_t^{T}$ using Eq.~(\ref{Eq.:S_final}), Eq.~(\ref{Eq.:T_final}) and Eq.~(\ref{Eq.:T_pooling})
    \STATE Calculate contrastive loss $\mathcal{L}^C_\mathcal{Q}$ and $\mathcal{L}^C_\mathcal{S}$ using Eq.~(\ref{eq:L^C_Q}) and Eq.~(\ref{eq:L^C_S})
    \STATE Calculate multi-label loss $\mathcal{L}_\text{list}$ using Eq.~(\ref{eq:multi_label})
    \STATE Calculate total loss $\mathcal{L}$ using Eq.~(\ref{eq:L_main})
    \STATE Update model parameter using Adam
\ENDWHILE
\RETURN $\theta$
\end{minipage}}
\end{algorithmic} 
\end{algorithm}

As shown in Figure~\ref{fig:method} (d), we capture high-order collaborative relationships between tools and align the cooperative interactions across two views using a cross-view contrastive loss.
Specifically, the representations of queries and scenes can be learned by optimizing the cross-view InfoNCE~\citep{gutmann2010noise,tang2024towards} loss:
\begin{align}
\mathcal{L}^C_\mathcal{Q} &= - \frac{1}{|\mathcal{Q}|} \sum_{q \in \mathcal{Q}} \log \frac{e^{\mathrm{sim}(\mathbf{e}_q^{S}, \mathbf{e}_q^{T}) / \tau}}{\sum_{q- \in \mathcal{Q}} e^{\mathrm{sim}(\mathbf{e}_q^{S}, {\mathbf{e}_{q-}^{T}}) / \tau}}, \label{eq:L^C_Q} \\
\mathcal{L}^C_\mathcal{S} &= - \frac{1}{|\mathcal{S}|} \sum_{s \in \mathcal{S}} \log \frac{e^{\mathrm{sim}(\mathbf{e}_s^{S}, \mathbf{e}_s^{T}) / \tau}}{\sum_{s- \in \mathcal{S}} e^{\mathrm{sim}(\mathbf{e}_s^{S}, {\mathbf{e}_{s-}^{T}}) / \tau}}, \label{eq:L^C_S}
\end{align}
where $\tau$ is the temperature parameter.

To ensure the complete retrieval of diverse tools from the full set of ground-truth tools, without favoring any particular tool, we design a list-wise multi-label loss as the main learning objective loss. 
Given a query $q$, the labeled training data is $\Gamma_q = \{\mathcal{T}_q=\{t_i,d_i\},y=\{y(q,t_i)\} |1 \leq i \leq L\}$, where $\mathcal{T}_q$ denotes a tool list with length $L$, comprising $N_q$ ground-truth tools and $L-N_q$ negative tools that are randomly sampled from the entire tool set.
$y(q,t_i)$ is the binary relevance label, taking a value of either 0 or 1, and the ideal scoring function should meet the following criteria:
\begin{equation}
p_q^t = \frac{\gamma(y(q,t))}{\sum_{t' \in \mathcal{T}_q} \gamma(y(q,t'))},
\end{equation}
where $p_q^t$ is the probability of selecting tool $t$.
$\gamma(y(q,t)) = 1$ if $y(q,t)=1$ and $\gamma(y(q,t)) = 0$ if $y(q,t)=0$. 

Similarly, given the predicted scores $\{\widehat{y}(q,t_1),\cdots,\widehat{y}(q,t_L)\}$, the probability of selecting tool $t$ can be derived:
\begin{equation}
\widehat{p_q^t} = \frac{\gamma(\widehat{y}(q,t))}{\sum_{t' \in \mathcal{T}_q} \gamma(\widehat{y}(q,t'))}.
\end{equation}

Therefore, the list-wise multi-label loss function minimizes the discrepancy between these two probability distributions:
\begin{equation}
\mathcal{L}_\text{list} = - \sum_{q \in \mathcal{Q}} \sum_{t \in \mathcal{T}_q} p_q^t \log \widehat{p_q^t} + (1 - p_q^t) \log(1 - \widehat{p_q^t}),
\label{eq:multi_label}
\end{equation}

Based on the multi-label loss $\mathcal{L}_\text{list}$ and the contrastive loss $\mathcal{L}^C_\mathcal{Q}$, the final loss $\mathcal{L}$ for our proposed COLT is formally defined as:
\begin{equation}
\mathcal{L} = \mathcal{L}_\text{list} + \lambda(\mathcal{L}^C_\mathcal{Q} + \mathcal{L}^C_\mathcal{S}),
\label{eq:L_main}
\end{equation}
where $\lambda$ is the co-efficient to balance the two losses. 

The learning algorithm of \ourmodel is summarized in Algorithm~\ref{alg:COLT}.

\section{Datasets}
To verify the effectiveness of \ourmodel, we utilize two datasets for multi-tool scenarios: ToolBench and a newly constructed dataset, \ourdata.
We randomly select $10\%$ of the entire dataset to serve as the test data. The statistics of the datasets after preprocessing are summarized in Table~\ref{tab:stat}.

\begin{table}[t]
\caption{Statistics of the experimental datasets. Tools/Query denotes the number of ground-truth tools for each query.}
\resizebox{0.9\columnwidth}{!}{
\begin{tabular}{cccccc}
\toprule
\multirow{2}{*}{Dataset} & \multicolumn{3}{c}{\# Query} & \multirow{2}{*}{\# Tool} & \multirow{2}{*}{\# Tools/Query}\\
\cmidrule(lr){2-4}
 & Training & Testing & Total \\
\midrule
ToolLens       & 16,893     & 1,877      & 18,770    & 464  & $1 \sim 3$ \\
ToolBench (I2) & 74,257     & 8,250      & 82,507    & 11,473  & $2 \sim 4$ \\
ToolBench (I3) & 21,361     & 2,373      & 23,734    & 1,419  & $2 \sim 4$ \\
\bottomrule
\end{tabular}}
\label{tab:stat}
\end{table}

\paratitle{ToolBench.}
ToolBench~\citep{qin2023toolllm} is a benchmark commonly used to evaluate the capability of LLMs in tool usage. 
In our experiments, we notice that its three subsets exhibit distinct characteristics. The first subset (I1) focuses on single-tool scenarios, which diverges from our emphasis on multi-tool tasks.
However, both the second subset (I2) and the third subset (I3) align with our focus on multi-tool tasks. Therefore, we chose I2 and I3 as the primary datasets for our experiments.

\paratitle{\ourdata.}
While existing datasets like ToolBench~\cite{qin2023toolllm} and TOOLE~\cite{huang2023metatool} provide multi-tool scenarios, they present limitations. TOOLE encompasses only 497 queries, and ToolBench's dataset construction, which involves providing complete tool descriptions to ChatGPT, results in verbose and semantically direct queries. These do not accurately reflect the brief and often multifaceted nature of real-world user queries. To address these shortcomings, we introduce \ourdata, crafted specifically for multi-tool scenarios.

As shown in Figure~\ref{fig:data}, the creation of \ourdata involves a novel five-step methodology:
\textbf{1) Tool Selection:} To create a high-quality tool dataset, we rigorously filter ToolBench, focusing on 464 available and directly callable tools relevant to everyday user queries, excluding those for authentication or testing.
\textbf{2) Scene Mining:} We prompt GPT-4 to generate potential scenes that are relevant to the detailed descriptions of the selected tools, and ensure their validity through human verification.
\textbf{3) Query Generation:} We then employ GPT-4 to generate queries based on the provided scene and the parameters required for tool calling. Notably, we avoid providing the complete tool description to GPT-4 to avoid the generated query being closely aligned with the tool.
\textbf{4) Tool Aggregation:} The queries generated in the aforementioned way are only relevant to a single tool. To enhance the relevance of queries across multiple tools, we reprocess them through GPT-4 to identify categories of tools that could be relevant, which are then aligned with our tool set through dense retrieval and manual verification.
\textbf{5) Query Rewriting:} Finally, we utilize GPT-4 to revise queries to incorporate all necessary parameters by providing it with both the initial query and a list of required parameters, thereby yielding concise yet intentionally multifaceted queries that better mimic real-world user interactions. 
It is worth noting that we incorporate a human verification process at each step to ensure data quality.

This comprehensive construction pipeline ensures \ourdata accurately simulates real-world tool retrieval dynamics. 
The resulting \ourdata dataset includes 18,770 queries and 464 tools, with each query being associated with $1 \sim 3$ verified tools.

\begin{table}[t]
\caption{Quality verification of \ourdata.}
\label{tab:quality_evalutation}%
  \resizebox{1.0\linewidth}{!}{
    \begin{tabular}{ccccc|ccc}
    \hline \hline
    \multicolumn{2}{c}{Evaluator} & \multicolumn{3}{c}{\ourdata vs. ToolBench} & \multicolumn{3}{c}{\ourdata vs. TOOLE} \\ \hline
    \multicolumn{8}{c}{\textbf{Whether the query is natural?}}  \\
    \hline
    \multicolumn{2}{c}{\multirow{2}{*}{GPT-4}} & \ourdata        & ToolBench  &Equal       & \ourdata    & TOOLE &Equal \\
    \multicolumn{2}{c}{} &68\%     & 14\%    & 18\%     & 44\%  & 36\%     & 20\%\\ 
    \hline
    \multicolumn{2}{c}{\multirow{2}{*}{Human}}  & \ourdata        & ToolBench  &Equal       & \ourdata    & TOOLE &Equal  \\
    \multicolumn{2}{c}{} &64\%     & 10\%    & 26\%     & 54\%  & 24\%     & 22\%\\
    \hline \hline
    \multicolumn{8}{c}{\textbf{Whether the user intent is multifaceted?}}  \\
    \hline
    \multicolumn{2}{c}{\multirow{2}{*}{GPT-4}}  & \ourdata        & ToolBench  &Equal       & \ourdata    & TOOLE &Equal  \\
    \multicolumn{2}{c}{} &62\%     & 14\%    & 24\%     & 50\%  & 24\%     & 26\%\\ 
    \hline
    \multicolumn{2}{c}{\multirow{2}{*}{Human}}  & \ourdata        & ToolBench  &Equal       & \ourdata    & TOOLE &Equal  \\
    \multicolumn{2}{c}{} &60\%     & 12\%    & 28\%     & 58\%  & 18\%     & 24\%\\
    \hline \hline
    \end{tabular}
    }
\end{table}

\paratitle{Discussion and Quality Verification.}
Unlike prior datasets, \ourdata uniquely focuses on creating natural, concise, and multifaceted queries to reflect real-world demands. To assess the quality of \ourdata, following previous works~\citep{Gao2023ConfuciusIT,liu2023geval,sottana-etal-2023-evaluation}, we employ GPT-4 as an evaluator and human evaluation where three well-educated doctor students are invited to evaluate 50 randomly sampled cases from \ourdata, ToolBench and TOOLE in the following two aspects:(1) Natural-query: whether the query is natural. (2) Multifaceted intentions: whether the user intent is multifaceted. The results are illustrated in Table ~\ref{tab:quality_evalutation}. In most cases, \ourdata outperforms ToolBench and TOOLE. Furthermore, using GPT-4 as the evaluator shows a high degree of consistency with human evaluation trends, which underscores the validity of employing GPT-4 as an evaluator.

\section{Experiments}
In this section, we first describe the experimental setups and then conduct an extensive evaluation and analysis of the proposed \ourmodel.
The source code and the proposed \ourdata dataset are publicly available at~\textcolor{blue}{\url{https://github.com/quchangle1/COLT}}.

\subsection{Experimental Setups}

\subsubsection{Evaluation Metrics.} 
Following the previous works~\cite{qin2023toolllm,Gao2023ConfuciusIT,zheng2024toolrerank}, we utilize the widely used retrieval metrics Recall@$K$ and NDCG@$K$ and report the metrics for $K \in \{3, 5\}$. However, as discussed in Section~\ref{introsection}, Recall and NDCG do not adequately fulfill the requirements of completeness that are crucial for effective tool retrieval. 
To further tailor our assessment to the specific challenges of tool retrieval tasks, we also introduce a new metric, \oureval@$K$. 
This metric is designed to measure whether the top-$K$ retrieved tools form a complete set with respect to the ground-truth set:
\[
\text{\oureval}\text{@$K$} = \frac{1}{|\mathcal{Q}|} \sum_{q=1}^{|\mathcal{Q}|} \mathbb{I}(\Phi_q \subseteq \Psi^K_q),
\]
where $\Phi_q$ is the set of ground-truth tools for query $q$, $\Psi^K_q$ represents the top-$K$ tools retrieved for query $q$, and $\mathbb{I}(\cdot)$ is an indicator function that returns 1 if the retrieval results include all ground-truth tools within the top-$K$ results for query $q$, and 0 otherwise.

\subsubsection{Baselines.} 
As our proposed \ourmodel is model-agnostic, we apply it to several representative PLM-based retrieval models~(as backbone models) to validate the effectiveness:

\textbf{ANCE}\citep{xiong2020approximate} uses a dual-encoder architecture with an asynchronously updated ANN index for training, enabling global selection of hard negatives. \textbf{TAS-B}\citep{hofstatter2021efficiently} is a bi-encoder that employs balanced margin sampling to ensure efficient query sampling from clusters per batch. \textbf{co-Condenser}\citep{gao2021unsupervised} uses a query-agnostic contrastive loss to cluster related text segments and distinguish unrelated ones. \textbf{Contriever}\citep{izacard2021unsupervised} leverages inverse cloze tasks, cropping for positive pair generation, and momentum contrastive learning to achieve state-of-the-art zero-shot retrieval performance.

In addition to PLM-based dense retrieval methods, we also compare with the classical lexical retrieval model BM25, widely used for tool retrieval as documented in  \cite{qin2023toolllm,Gao2023ConfuciusIT}.
\textbf{BM25}~\citep{robertson2009probabilistic} uses an inverted index to identify suitable tools based on exact term matching. 

\begin{table*}[t]
    \caption{Performance comparison of different tool retrieval methods on \ourdata and ToolBench datasets. ``$^\dagger$'' denotes the best results for each column. The term ``Zero-shot'' refers to the performance of dense retrieval models without any training. ``+Fine-tune'' indicates that retrieval models are fine-tuned on \ourdata and ToolBench datasets. ``+\ourmodel (Ours)'' indicates that dense retrieval backbones are equipped with our proposed method.  R@$K$, N@$K$, and C@$K$ are short for Recall@$K$, NDCG@$K$ and \oureval@$K$, respectively.}
    \centering
    \resizebox{1\textwidth}{!}{
    \begin{tabular}{lccccccccccccccccccc}
        \toprule
        \multirow{2}{*}{Backbone} & \multirow{2}{*}{Framework} & \multicolumn{6}{c}{\textbf{\ourdata}} & \multicolumn{6}{c}{\textbf{ToolBench (I2)}} & \multicolumn{6}{c}{\textbf{ToolBench (I3)}}\\ 
        \cmidrule(lr){3-8} \cmidrule(lr){9-14} \cmidrule(lr){15-20}
        & & R@3 & R@5 & N@3 & N@5 & C@3 & C@5 & R@3 & R@5 & N@3 & N@5 & C@3 & C@5 & R@3 & R@5 & N@3 & N@5 & C@3 & C@5\\ 
        \midrule
        {BM25} & - & 21.58 & 26.88 & 23.19 & 26.09 & 3.89 & 6.13 & 17.06 & 21.38 & 17.83 & 19.88 & 2.39 & 4.37 & 29.33 & 35.88 & 32.20 & 35.08 & 5.52 & 9.78\\
        \midrule[0.3pt]
        \multirow{3}{*}{ANCE} & Zero-shot & 20.82 & 26.56 & 21.45 & 24.57 & 5.06 & 7.46 & 20.82 & 26.56 & 21.45 & 24.57 & 5.06 & 7.46 & 21.55 & 26.38 & 23.44 & 25.60 & 2.44 & 4.59 \\
        & +Fine-tune & 80.62 & 94.17 & 82.35 & 90.15 & 54.23 & 85.83 & 58.58 & 67.20 & 58.58 & 63.75 & 26.46 & 42.80 & 65.11 & 76.63 & 69.27 & 74.14 & 34.68 & 53.64 \\
        & +\ourmodel (Ours) & \textbf{92.15} & \textbf{97.78}$^\dagger$ & \textbf{92.78} & \textbf{96.10} & \textbf{80.50} & \textbf{94.40} & \textbf{70.76} & \textbf{80.59} & \textbf{73.64} & \textbf{77.98} & \textbf{45.10} & \textbf{62.93} & \textbf{73.37} & \textbf{83.97} & \textbf{77.95} & \textbf{82.14} & \textbf{46.01} & \textbf{66.41}\\
        \midrule[0.3pt]
        \multirow{3}{*}{TAS-B} & Zero-shot & 19.10 & 23.71 & 19.81 & 22.33 & 5.17 & 7.14 & 19.10 & 23.71 & 19.81 & 22.33 & 5.17 & 7.14 & 25.32 & 31.15 & 27.80 & 30.36 & 3.84 & 6.40 \\
        & +Fine-tune & 81.26 & 94.06 & 82.54 & 89.94 & 54.66 & 85.72 & 62.78 & 67.49 & 58.96 & 64.21 & 26.74 & 43.66 & 66.04 & 77.64 & 70.41 & 75.34 & 35.69 & 55.75 \\
        & +\ourmodel (Ours) & \textbf{91.49} & \textbf{96.91} & \textbf{92.48} & \textbf{95.63} & \textbf{79.00} & \textbf{92.22} & \textbf{71.64} & \textbf{81.12} & \textbf{74.60} & \textbf{78.74} & \textbf{46.77} & \textbf{64.38} & \textbf{74.49} & \textbf{84.58} & \textbf{79.03} & \textbf{82.95} & \textbf{48.16} & \textbf{68.35}\\
        \midrule[0.3pt]
        \multirow{3}{*}{coCondensor} & Zero-shot & 15.33 & 19.37 & 16.15 & 18.32 & 3.02 & 5.33 & 15.33 & 19.37 & 16.15 & 18.32 & 3.02 & 5.30 & 20.80 & 25.24 & 23.21 & 25.10 & 2.07 & 3.75 \\
        & +Fine-tune & 82.37 & 94.69 & 83.90 & 91.06 & 56.37 & 86.73 & 57.70 & 69.46 & 60.80 & 66.07 & 28.78 & 46.06 & 66.97 & 79.30 & 71.20 & 76.50 & 37.08 & 58.66 \\
        & +\ourmodel (Ours) & \textbf{92.65} & \textbf{97.78}$^\dagger$ & \textbf{93.16} & \textbf{96.17} & \textbf{82.25} & \textbf{94.56}$^\dagger$ & \textbf{73.91} & \textbf{83.47} & \textbf{76.75} & \textbf{80.87} & \textbf{49.15} & \textbf{67.75} & \textbf{75.48} & \textbf{84.97} & \textbf{80.00} & \textbf{83.55} & \textbf{49.17} & \textbf{68.64}$^\dagger$\\
        \midrule[0.3pt]
        \multirow{3}{*}{Contriever} & Zero-shot & 25.67 & 31.15 & 26.96 & 29.95 & 7.46 & 9.80 & 25.67 & 31.15 & 26.96 & 29.95 & 7.46 & 9.80 & 31.37 & 38.60 & 34.13 & 37.37 & 6.03 & 11.42 \\
        & +Fine-tune & 83.58 & 95.17 & 84.98 & 91.69 & 59.46 & 88.65 & 58.89 & 70.75 & 62.11 & 67.42 & 29.77 & 48.31 & 68.58 & 80.05 & 72.86 & 77.69 & 39.70 & 60.89 \\
        & +\ourmodel (Ours) & \textbf{93.64}$^\dagger$ & \textbf{97.75} & \textbf{94.53}$^\dagger$ & \textbf{96.91}$^\dagger$ & \textbf{84.55}$^\dagger$ & \textbf{94.08} & \textbf{75.72}$^\dagger$ & \textbf{85.03}$^\dagger$ & \textbf{78.57}$^\dagger$ & \textbf{82.54}$^\dagger$ & \textbf{51.97}$^\dagger$ & \textbf{70.10}$^\dagger$ & \textbf{76.63}$^\dagger$ & \textbf{85.50}$^\dagger$ & \textbf{81.21}$^\dagger$ & \textbf{84.18}$^\dagger$ & \textbf{52.00}$^\dagger$ & \textbf{68.47}\\
        \bottomrule
    \end{tabular}
    }
    \label{tab:main}
\end{table*}

\subsubsection{Implementation Details.} We utilize the BEIR~\citep{thakur2021beir} framework for dense retrieval baselines, setting the training epochs to $5$ with the learning rate of $2e\text{--}5$, weight decay of $0.01$, and using the AdamW optimizer. Our model-agnostic approach directly applies dense retrieval for semantic learning. During collaborative learning, we set the batch size as 2048 and carefully tune the learning rate among $\{1e\text{--}3, 5e\text{--}3,1e\text{--}4,5e\text{--}4,1e\text{--}5\}$, the weight decay among  $\{1e\text{--}5, 1e\text{--}6, 1e\text{--}7\}$, as well as the layer number $I$ among $\{1, 2, 3\}$.

\subsection{Experimental Results}

\subsubsection{Retrieval Performance.}
Table~\ref{tab:main} presents the results of different tool retrieval methods on \ourdata, ToolBench~(I2 and I3). From the results, we have the following observations and conclusions:

We can observe that traditional dense retrieval models perform poorly in zero-shot scenarios, even inferior to that of BM25. This indicates that these models may not be well-suited for tool retrieval tasks. 
Conversely, the BM25 model significantly lags behind fine-tuned PLM-based dense retrieval methods, underscoring the superior capability of the latter in leveraging contextual information for more effective tool retrieval.
Despite this advantage, PLM-based methods fall short in the \Oureval metric, which is specifically designed for evaluating completeness in tool retrieval scenarios. 
This suggests that while effective for general retrieval tasks, PLM-based methods may not fully meet the unique demands of tool retrieval.

All base models equipped with \ourmodel exhibit significant performance gains across all metrics on all three datasets, particularly in the \oureval@3 metric. 
These improvements demonstrate the effectiveness of \ourmodel, which can be attributed to the fact that \ourmodel adopts a two-stage learning framework with semantic learning followed by collaborative learning.
In this way, COLT can capture the intricate collaborative relationships between tools, resulting in effectively retrieving a complete tool set.

\subsubsection{Downstream Tool Learning Performance.}
To verify that improvements of \ourmodel in tool retrieval truly enhance downstream tool learning applications, we conduct a validation study using the pairwise comparison method~\cite{Dai_2023,sun2023instruction,liang2023holistic}. 
We randomly select 100 queries from the test set of \ourdata and use various retrieval models to retrieve the top-3 tools for each query.
Then we utilize GPT-4 as an evaluator, examining the responses generated with different retrieved tools across four dimensions: Coherence, Relevance, Comprehensiveness, and Overall. 
Specifically, the user query and a pair of responses are utilized as prompts to guide GPT-4 in determining the superior response. 
Additionally, we also consider that LLMs may respond differently to the order in which text is presented in the prompt~\citep{lu2022fantastically, tang2023middle, hou2024large,liu2023lost}. 
So each comparison is conducted twice with reversed response order to mitigate potential biases from text order, ensuring a more reliable assessment.

\begin{table}[t]
\caption{
Elo ratings for different models w.r.t. ``Coherence'', ``Relevance'', ``Comprehensiveness'' and  ``Overall'' evaluated by GPT-4 on \ourdata dataset.
}
\centering
\scriptsize
\setlength{\tabcolsep}{2pt}

\begin{tabular}{lcp{6.8mm}cp{6.8mm}cp{6.8mm}cp{6.8mm}}
\toprule
                    & \multicolumn{8}{c}{Evaluation Aspects}
                    \\ \cmidrule(rl){2-9}
                    & \multicolumn{2}{c}{Coherence}           
                    & \multicolumn{2}{c}{Relevance}         
                    & \multicolumn{2}{c}{Comprehensiveness}   
                    & \multicolumn{2}{c}{Overall}  \\ \midrule
{BM25}        & \phantom{0}848           & \chart{848}{9}{cyan} & \phantom{0}845           & \chart{845}{8}{cyan} & \phantom{0}860           & \chart{860}{10}{cyan}  & \phantom{0}780           & \chart{780}{5}{cyan}\\

{ANCE} & \phantom{0}934 & \chart{934}{23}{cyan}  & \phantom{0}936           & \chart{936}{23}{cyan} & \phantom{0}946           & \chart{946}{26}{cyan} & 1016           & \chart{1016}{48}{cyan}\\ 

{TAS-B} & \phantom{0}995 & \chart{995}{41}{cyan}  & \phantom{0}991           & \chart{991}{39}{cyan} & \phantom{0}988           & \chart{988}{39}{cyan} & 1028           & \chart{1028}{60}{cyan}\\

{coCondensor} & 1031 & \chart{1031}{54}{cyan}  & 1036           & \chart{1036}{56}{cyan} & 1041           & \chart{1041}{58}{cyan} & 1035           & \chart{1035}{55}{cyan}\\
{Contriever} & 1076 & \chart{1076}{73}{cyan}  & 1082           & \chart{1082}{76}{cyan} & 1044           & \chart{1044}{59}{cyan} & 1046           & \chart{1046}{60}{cyan}\\
\midrule
{COLT (Ours)} & 1116 & \chart{1116}{93}{cyan}  & 1110           & \chart{1110}{90}{cyan} & 1121           & \chart{1121}{95}{cyan} & 1096           & \chart{1096}{82}{cyan}\\
\bottomrule
\end{tabular}
\label{tab:downstream evaluation}
\end{table}

We establish a tournament-style competition using the Elo ratings system, which is widely employed in chess and other two-player games to measure the relative skill levels of the players~\citep{dettmers2023qlora,wu2023style}. Following previous works~\citep{vicuna2023}, we start with a score of $1,000$ and set $K$-factor to $32$. Additionally, to minimize the impact of match sequences on Elo scores, we conduct these computations $10,000$ times using various random seeds to control for ordering effects.

The results in Table~\ref{tab:downstream evaluation} show that superior tool retrieval models can significantly improve downstream tool learning performance. Moreover, responses generated with the tools retrieved from \ourmodel notably outperform those from other methods, achieving the highest Elo ratings in all four assessed dimensions. These results highlight the pivotal role of effective tool retrieval in tool learning applications and further confirm the superiority of \ourmodel.

\subsection{Further Analysis}
Next, we delve into investigating the effectiveness of \ourmodel. We report the experimental results on the \ourdata and ToolBench~(I3) datasets, observing similar trends on ToolBench~(I2). Recall@|N| and \oureval@|N| are adopted as evaluation metrics, with |N| representing the count of ground-truth tools suitable for each query.

\begin{table}[t]
\caption{Ablation study of the proposed \ourmodel.}
	\small
	\centering
        \resizebox{0.9\linewidth}{!}{
	\begin{tabular}{lcccc}
		\toprule
		\multirow{2.5}{*}{Methods} & \multicolumn{2}{c}{\textbf{\ourdata}} & \multicolumn{2}{c}{\textbf{ToolBench}} \\ 
		\cmidrule(lr){2-3} \cmidrule(lr){4-5}
		& R@|N| & C@|N| & R@|N| & C@|N|\\ 
		\midrule
		\textbf{ANCE+\ourmodel (Ours)}           & \textbf{91.08} & \textbf{78.36} & \textbf{72.22} & \textbf{44.28}\\
            \quad \wo semantic learning  &36.49 & 6.84 & 21.92 & 1.60 \\
            \quad \wo collaborative learning &77.36 & 49.01 & 62.39 & 30.12 \\
            \quad \wo list-wise learing &79.94 & 52.68 & 66.02 & 35.82 \\
            \quad \wo contrastive learning &85.63 & 63.87 & 66.57 & 34.55 \\
            \midrule
        \textbf{TAS-B+\ourmodel (Ours)}           & \textbf{90.29} & \textbf{77.73} & \textbf{72.84} & \textbf{45.46}\\
            \quad \wo semantic learning  &38.49 & 9.16 & 32.16 & 5.47 \\
            \quad \wo collaborative learning &76.86 & 47.83 & 63.61 & 31.73 \\
            \quad \wo list-wise learning &79.89 & 52.25 & 66.91 & 37.27 \\
            \quad \wo contrastive learning &84.86 & 62.65 & 67.66 & 36.36 \\
            \midrule
        \textbf{coCondensor+\ourmodel (Ours)}           & \textbf{91.49} & \textbf{79.86} & \textbf{74.00} & \textbf{47.49}\\
            \quad \wo semantic learning  &30.38 & 5.54 & 25.07 & 2.27 \\
            \quad \wo collaborative learning &78.83 & 50.61 & 64.38 & 33.08 \\
            \quad \wo list-wise learning &81.42 & 54.16 & 69.18 & 40.67 \\
            \quad \wo contrastive learning &86.78 & 67.07 & 68.92 & 37.80 \\
            \midrule
		\textbf{Contriever+\ourmodel (Ours)}           & \textbf{92.76} & \textbf{82.95} & \textbf{75.40} & \textbf{49.81}\\
            \quad \wo semantic learning  &65.21 & 30.90 & 53.33 & 19.63 \\
            \quad \wo collaborative learning &80.60 & 54.44 & 68.20 & 36.91 \\
            \quad \wo list-wise learning &81.49 & 54.93 & 71.80 & 46.07 \\
            \quad \wo contrastive learning &84.58 & 60.52 & 69.46 & 39.02 \\
		\bottomrule
	\end{tabular}
	}
	\label{tab:ab}
\end{table}

\subsubsection{Ablation Study.} 
We conduct ablation studies to assess the impact of various components within our \ourmodel. The results presented in Table~\ref{tab:ab}, highlight the significance of each element:

\textbf{\wo semantic learning} denotes an off-the-shelf PLM is directly employed to get the initial representation for the subsequent collaborative learning stage without semantic learning on the given dataset in Section~\ref{semantic learning}. The absence of semantic learning significantly diminishes performance, confirming its essential role in aligning the representations of tools and queries as the basic for the following collaborative learning. Notably, the omission of semantic learning elements markedly reduces performance across other models more than with Contriever.
This highlights the superior ability of Contriever in zero-shot learning scenarios compared to the other models.

\textbf{\wo collaborative learning} is a variant where the collaborative learning state is omitted (\ie only semantic learning). The significant decline in performance in this variant further supports the effectiveness of \ourmodel in capturing the high-order relationships between tools through graph collaborative learning, thereby achieving comprehensive tool retrieval. 

\textbf{\wo list-wise learning} refers to a variant that optimizes using pair-wise loss in place of the list-wise loss defined in Eq.~\eqref{eq:multi_label}.
This substitution results in a significant drop in performance, highlighting that compared to pairwise loss, list-wise loss optimizes the tools in the same scenario as a whole entity, proving more effective in focusing on completeness.

\textbf{\wo contrastive learning} refers to a variant that optimizes without the contrastive loss defined in Eq.~\eqref{eq:L^C_Q} and~\eqref{eq:L^C_S}; 
This omission also leads to a noticeable performance drop, emphasizing the benefits of introducing contrastive learning to achieve better representation for queries and tools within a dual-view learning framework.
Additionally, our analysis reveals that contrastive learning is particularly crucial for Contriever, as its absence results in performance lagging behind the other models. This also indicates that the importance of contrastive learning varies across different backbones.
\begin{figure}[t]
    \centering
    \subfigtopskip=2pt 
	\subfigbottomskip=2pt 
	\subfigcapskip=-5pt 
    \subfigure[\ourdata]{
    \includegraphics[width=0.46\linewidth]{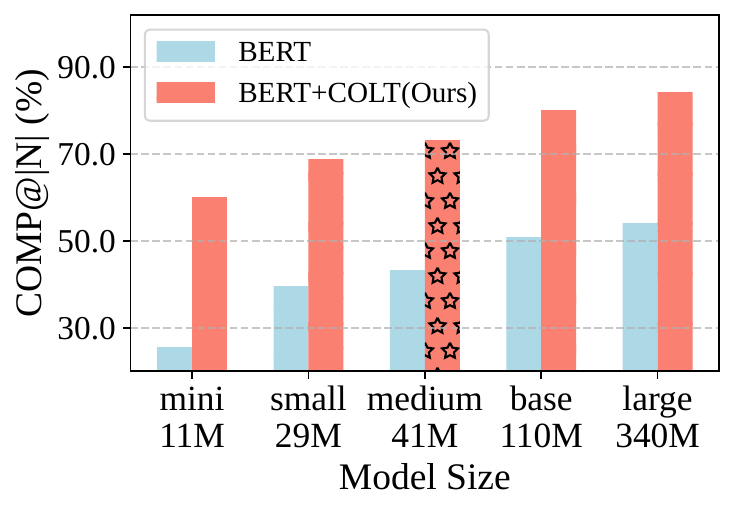}
    }
    \subfigure[ToolBench]{
    \includegraphics[width=0.46\linewidth]{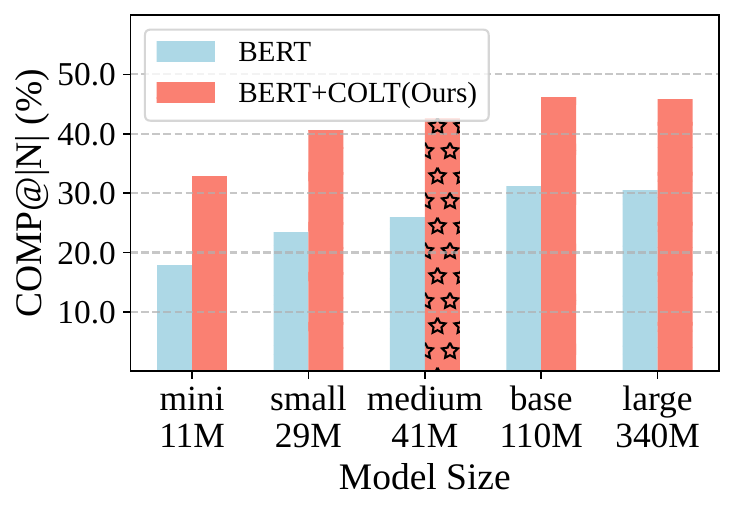}
    }
    \caption{Comparison of different model sizes of PLM.}
    \vspace{-1em}
    \label{fig:analy_plmsize}
\end{figure}

\begin{figure*}[t]
    \centering
    \subfigtopskip=2pt 
	\subfigbottomskip=1pt 
	\subfigcapskip=-8pt 
    \subfigure[\ourdata]{
    \includegraphics[width=0.23\textwidth]{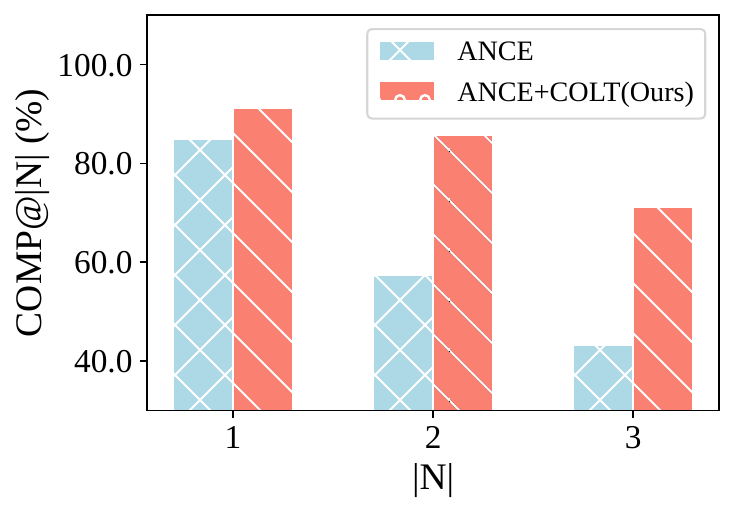}
    \hspace{1mm}
    \includegraphics[width=0.23\textwidth]{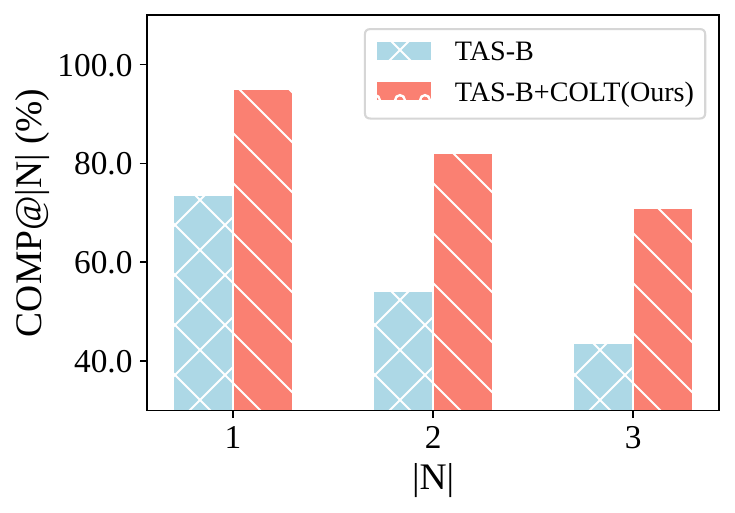}
    \hspace{1mm}
    \includegraphics[width=0.23\textwidth]{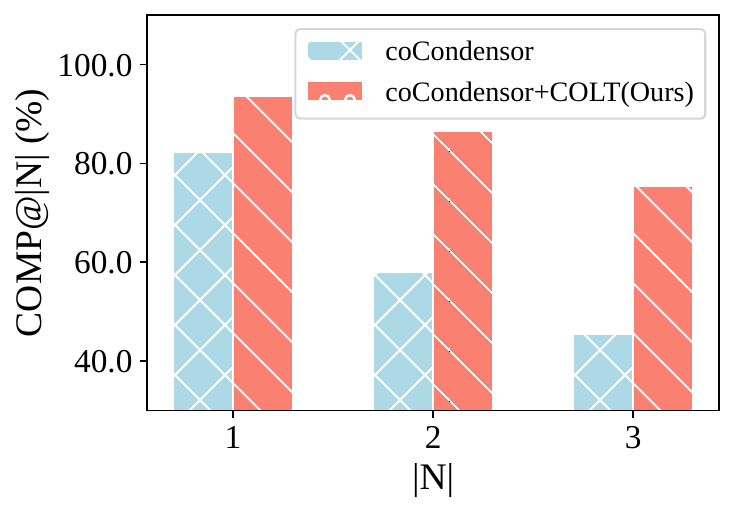}
    \hspace{1mm}
    \includegraphics[width=0.23\textwidth]{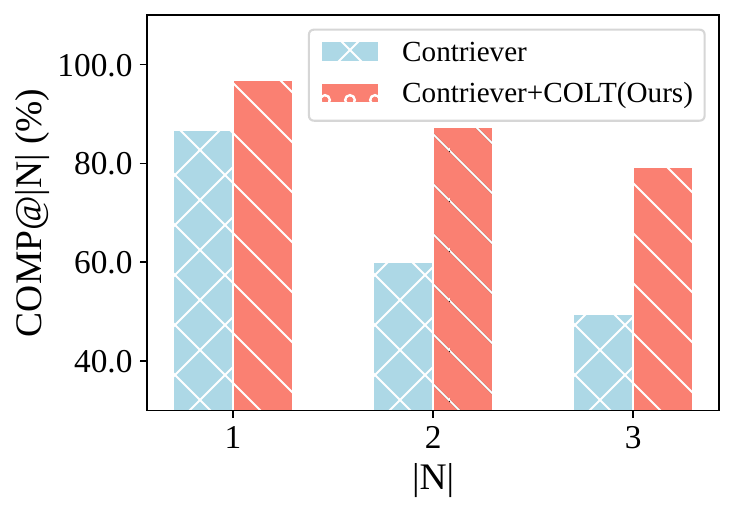}
    }
    \subfigure[ToolBench]{
    \includegraphics[width=0.23\textwidth]{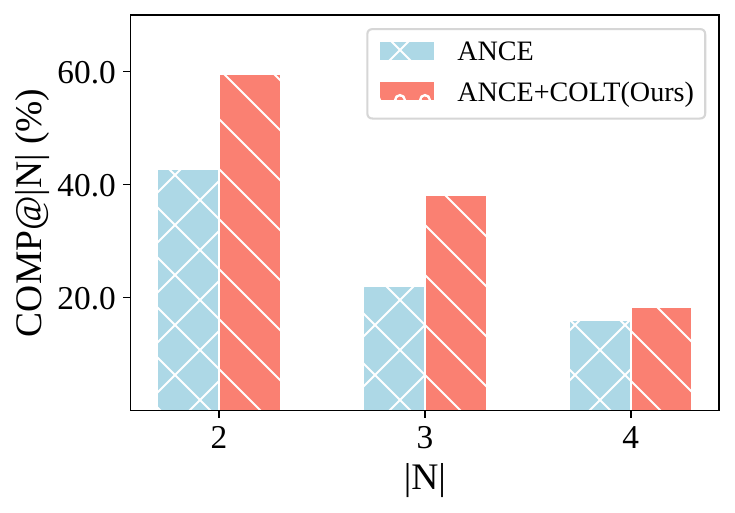}
    \hspace{1mm}
    \includegraphics[width=0.23\textwidth]{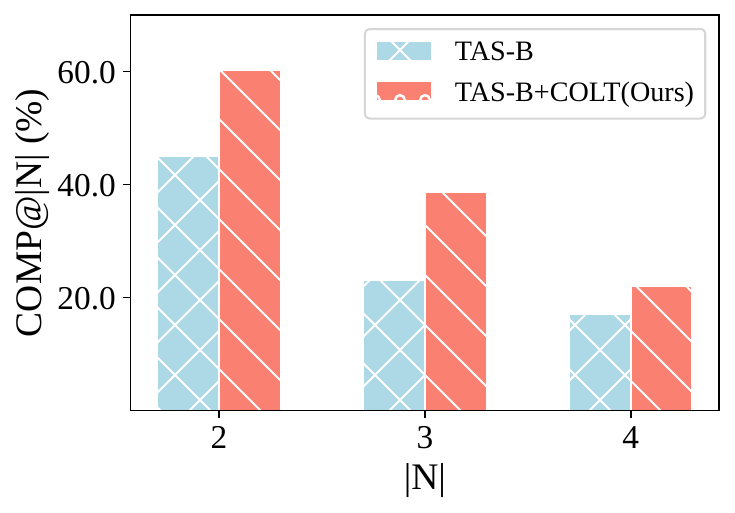}
    \hspace{1mm}
    \includegraphics[width=0.23\textwidth]{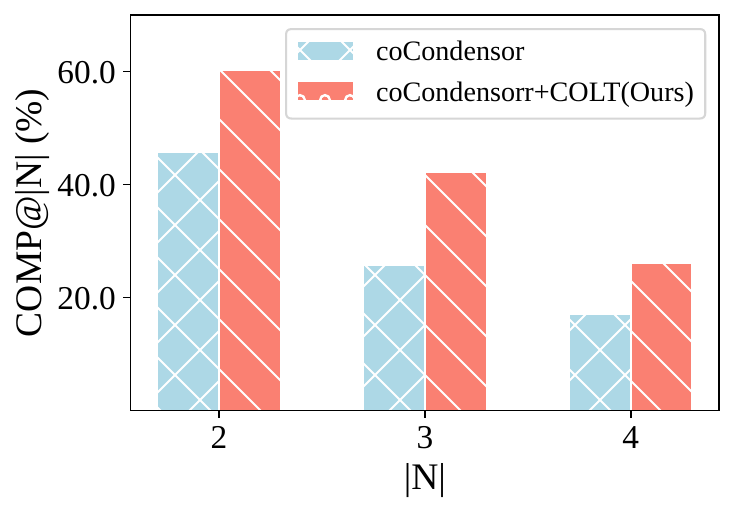}
    \hspace{1mm}
    \includegraphics[width=0.23\textwidth]{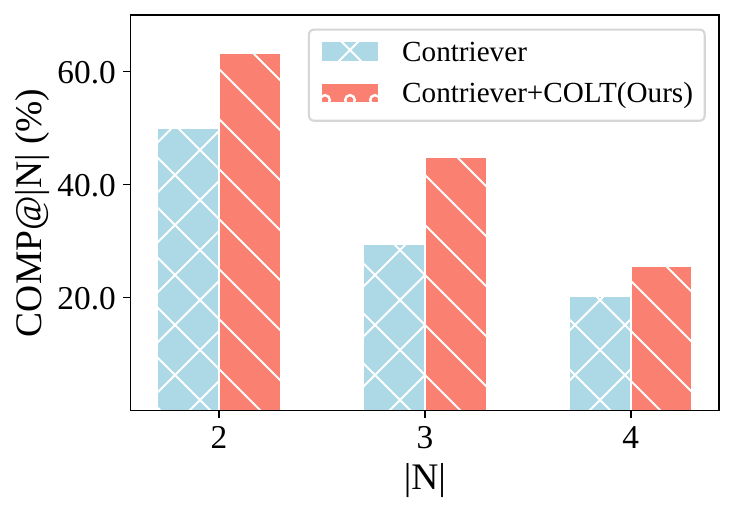}
    }    
    \vspace{-0.5em}
    \caption{Performance comparison regarding different sizes of ground-truth tool sets.}
    \vspace{-0.4em}
    \label{fig:analy_toolsize}
\end{figure*}

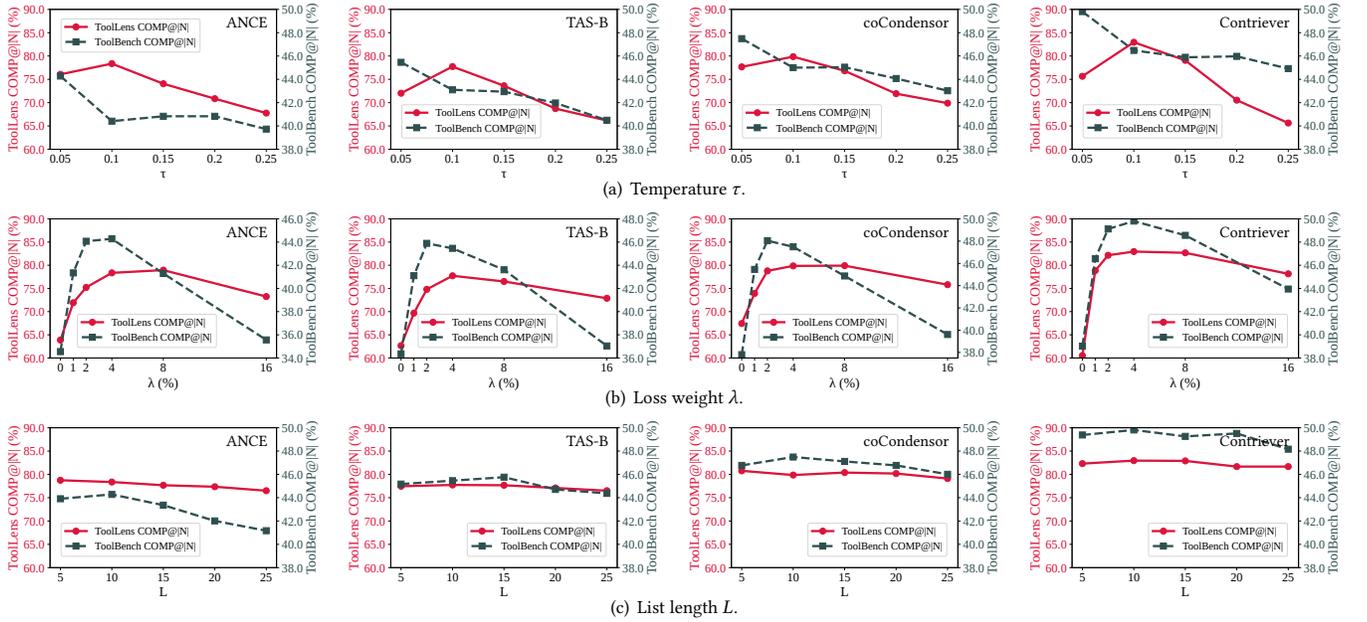
\begin{figure*}[t]
	\centering
    \subfigtopskip=2pt 
	\subfigbottomskip=1pt 
	\subfigcapskip=-8pt 
	\subfigure[Temperature $\tau$.]{\label{fig:temp}
		\begin{tikzpicture}
			\node[anchor=south west,inner sep=0] (image) at (0,0) {\includegraphics[width=0.24\textwidth]{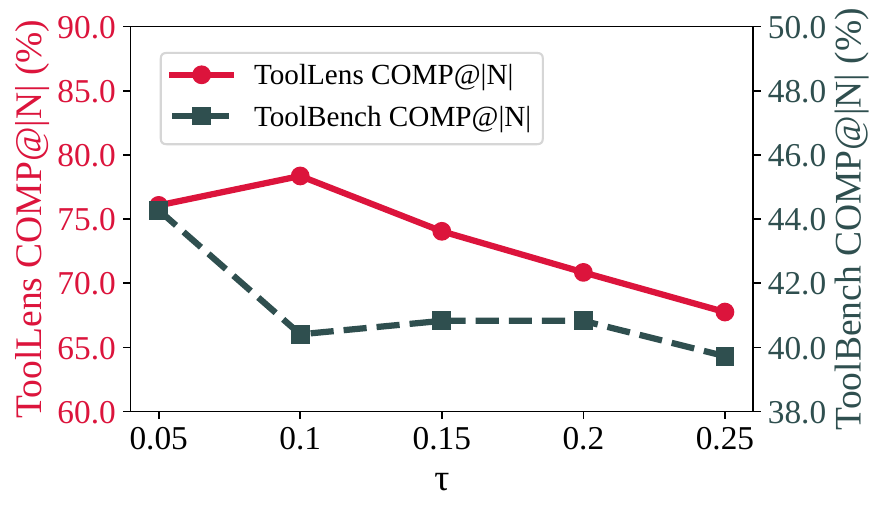}};
            \begin{scope}[x={(image.south east)},y={(image.north west)}]
            \node[anchor=north east, color=black] at (0.85,0.95) {\fontsize{6}{8}\selectfont ANCE};
            \end{scope}
		\end{tikzpicture}
        \hspace{1mm}
        \begin{tikzpicture}
			\node[anchor=south west,inner sep=0] (image) at (0,0) {\includegraphics[width=0.24\textwidth]{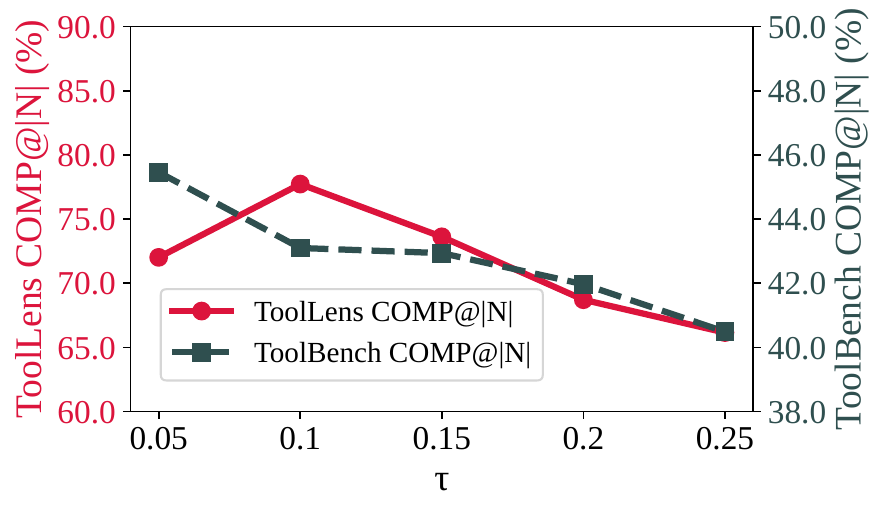}};
            \begin{scope}[x={(image.south east)},y={(image.north west)}]
            \node[anchor=north east, color=black] at (0.85,0.95) {\fontsize{6}{8}\selectfont TAS-B};
            \end{scope}
		\end{tikzpicture}
        \hspace{1mm}
        \begin{tikzpicture}
			\node[anchor=south west,inner sep=0] (image) at (0,0) {\includegraphics[width=0.24\textwidth]{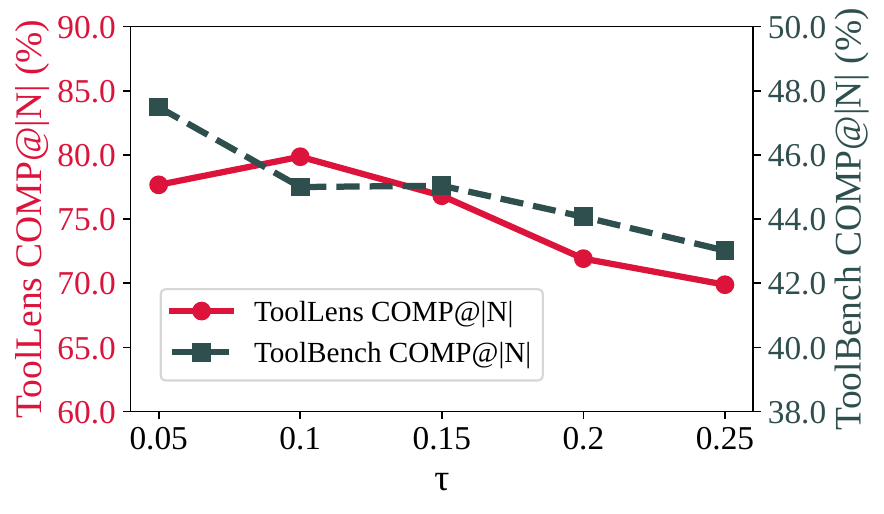}};
            \begin{scope}[x={(image.south east)},y={(image.north west)}]
            \node[anchor=north east, color=black] at (0.85,0.95) {\fontsize{6}{8}\selectfont coCondensor};
            \end{scope}
		\end{tikzpicture}
        \hspace{1mm}
        \begin{tikzpicture}
			\node[anchor=south west,inner sep=0] (image) at (0,0) {\includegraphics[width=0.24\textwidth]{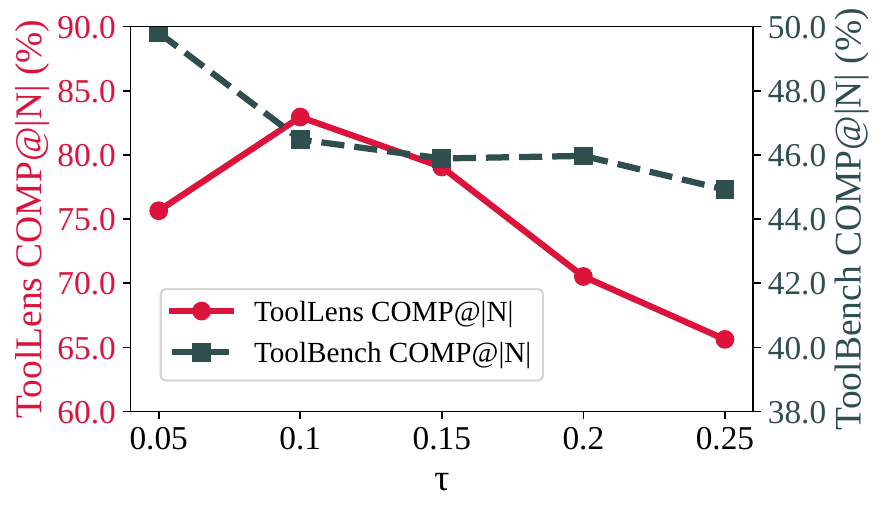}};
            \begin{scope}[x={(image.south east)},y={(image.north west)}]
            \node[anchor=north east, color=black] at (0.85,0.95) {\fontsize{6}{8}\selectfont Contriever};
            \end{scope}
		\end{tikzpicture}
	}
	\subfigure[Loss weight $\lambda$.]{\label{fig:lossweight}
		\begin{tikzpicture}
			\node[anchor=south west,inner sep=0] (image) at (0,0) {\includegraphics[width=0.24\textwidth]{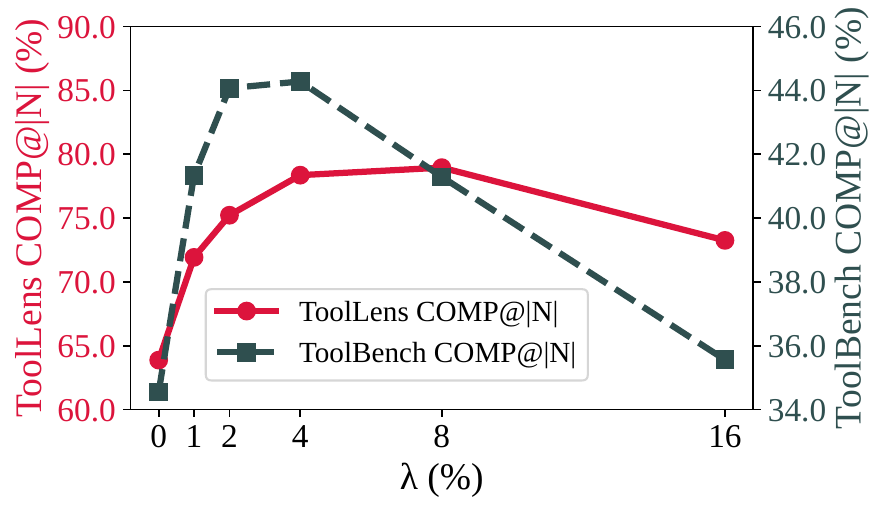}};
            \begin{scope}[x={(image.south east)},y={(image.north west)}]
            \node[anchor=north east, color=black] at (0.85,0.95) {\fontsize{6}{8}\selectfont ANCE};
            \end{scope}
		\end{tikzpicture}
        \hspace{1mm}
        \begin{tikzpicture}
			\node[anchor=south west,inner sep=0] (image) at (0,0) {\includegraphics[width=0.24\textwidth]{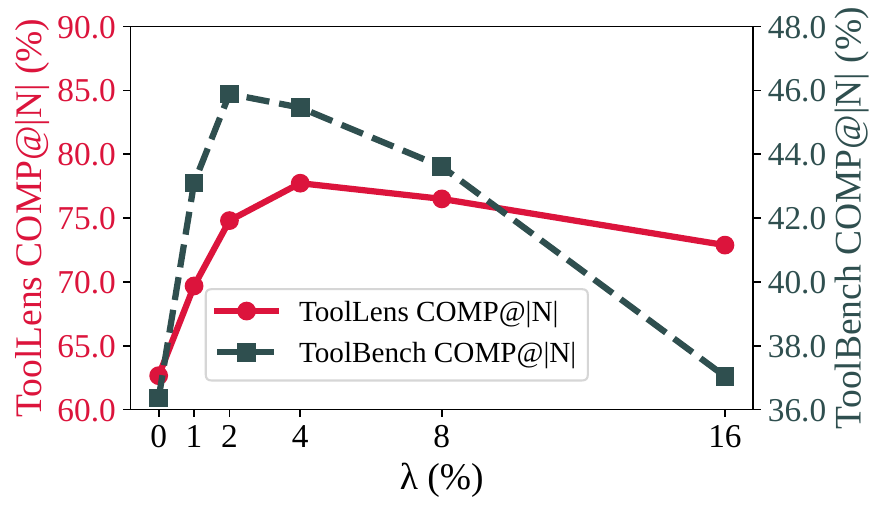}};
			\begin{scope}[x={(image.south east)},y={(image.north west)}]
            \node[anchor=north east, color=black] at (0.85,0.95) {\fontsize{6}{8}\selectfont TAS-B};
            \end{scope}
		\end{tikzpicture}
        \hspace{1mm}
        \begin{tikzpicture}
			\node[anchor=south west,inner sep=0] (image) at (0,0) {\includegraphics[width=0.24\textwidth]{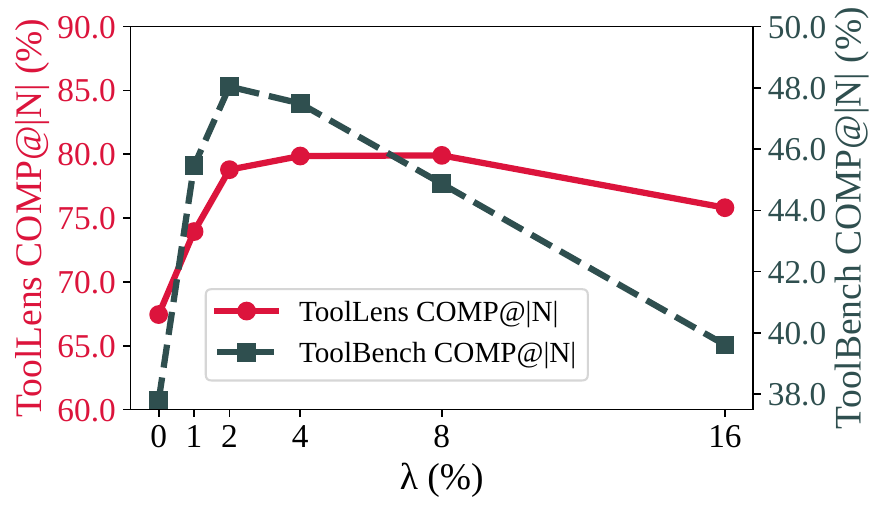}};
			\begin{scope}[x={(image.south east)},y={(image.north west)}]
            \node[anchor=north east, color=black] at (0.85,0.95) {\fontsize{6}{8}\selectfont coCondensor};
            \end{scope}
		\end{tikzpicture}
        \hspace{1mm}
        \begin{tikzpicture}
			\node[anchor=south west,inner sep=0] (image) at (0,0) {\includegraphics[width=0.24\textwidth]{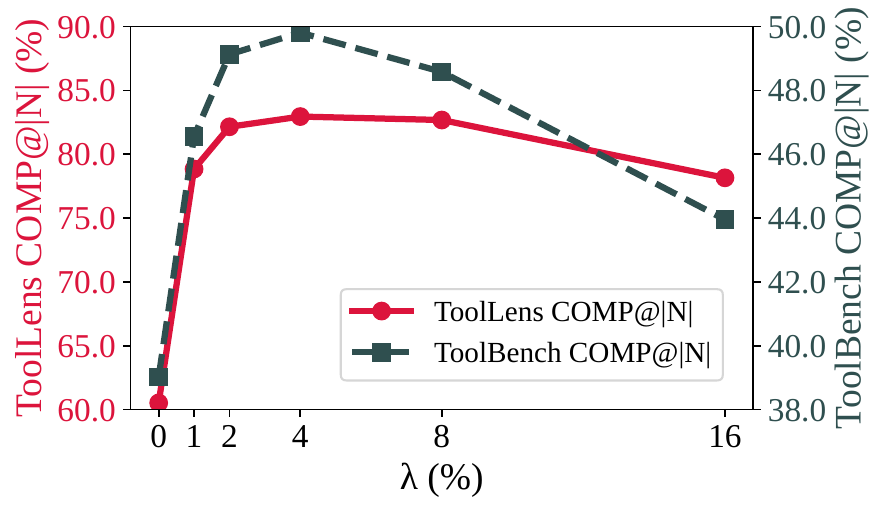}};
			\begin{scope}[x={(image.south east)},y={(image.north west)}]
            \node[anchor=north east, color=black] at (0.85,0.95) {\fontsize{6}{8}\selectfont Contriever};
            \end{scope}
		\end{tikzpicture}
	}
	\subfigure[List length $L$.]{\label{fig:inputlength}
		\begin{tikzpicture}
			\node[anchor=south west,inner sep=0] (image) at (0,0) {\includegraphics[width=0.24\textwidth]{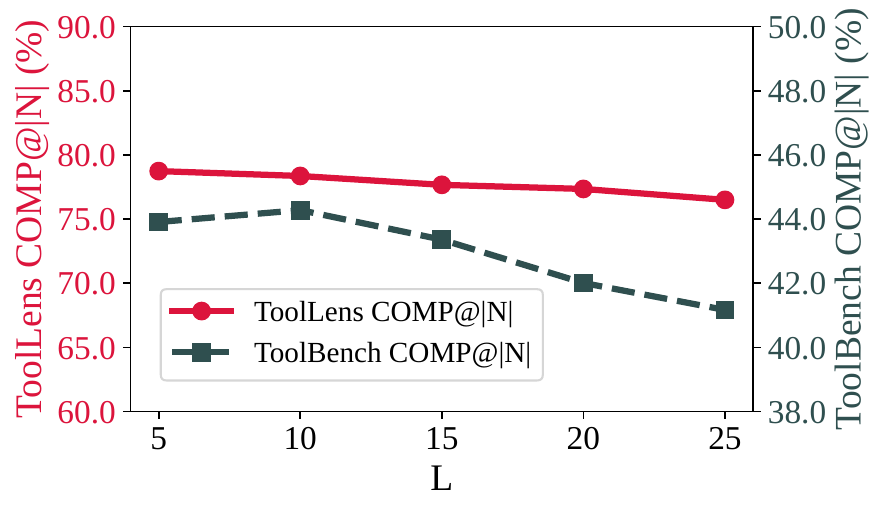}};
			\begin{scope}[x={(image.south east)},y={(image.north west)}]
            \node[anchor=north east, color=black] at (0.85,0.95) {\fontsize{6}{8}\selectfont ANCE};
            \end{scope}
		\end{tikzpicture}
        \hspace{1mm}
        \begin{tikzpicture}
			\node[anchor=south west,inner sep=0] (image) at (0,0) {\includegraphics[width=0.24\textwidth]{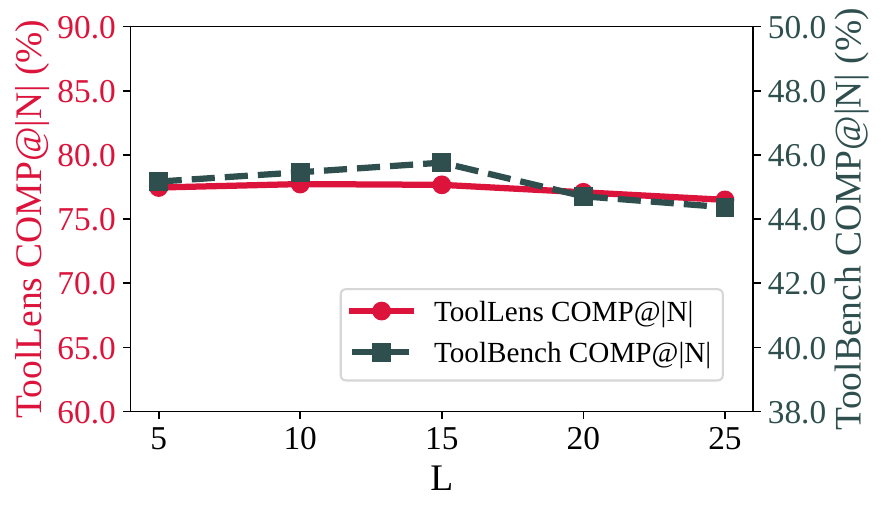}};
			\begin{scope}[x={(image.south east)},y={(image.north west)}]
            \node[anchor=north east, color=black] at (0.85,0.95) {\fontsize{6}{8}\selectfont TAS-B};
            \end{scope}
		\end{tikzpicture}
        \hspace{1mm}    
        \begin{tikzpicture}
			\node[anchor=south west,inner sep=0] (image) at (0,0) {\includegraphics[width=0.24\textwidth]{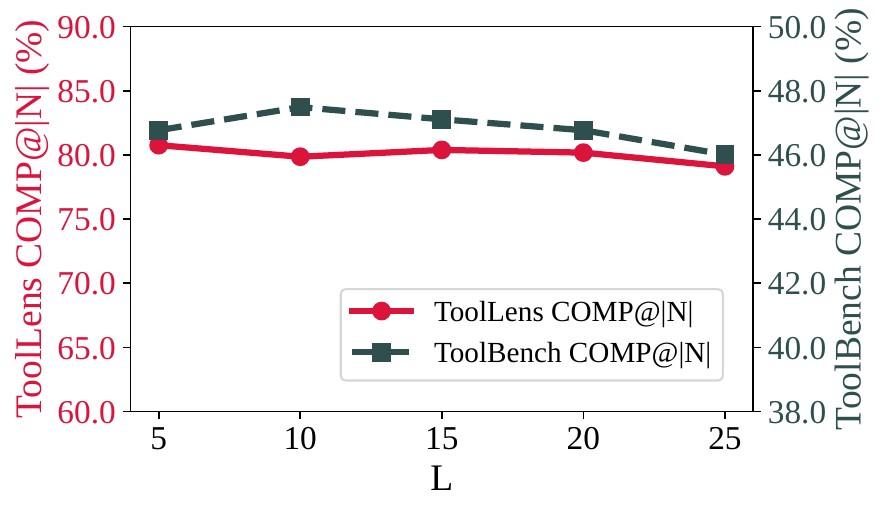}};
			\begin{scope}[x={(image.south east)},y={(image.north west)}]
            \node[anchor=north east, color=black] at (0.85,0.95) {\fontsize{6}{8}\selectfont coCondensor};
            \end{scope}
		\end{tikzpicture}
        \hspace{1mm}    
        \begin{tikzpicture}
			\node[anchor=south west,inner sep=0] (image) at (0,0) {\includegraphics[width=0.24\textwidth]{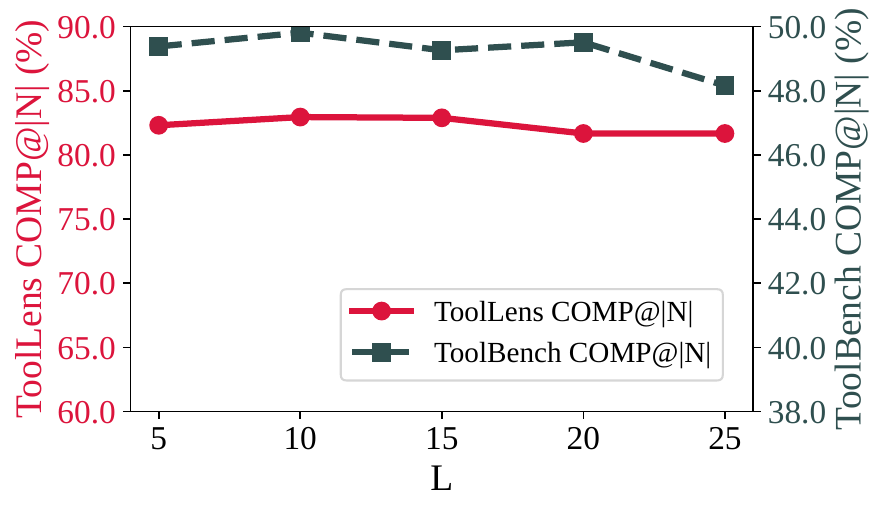}};
			\begin{scope}[x={(image.south east)},y={(image.north west)}]
            \node[anchor=north east, color=black] at (0.85,0.95) {\fontsize{6}{8}\selectfont Contriever};
            \end{scope}
		\end{tikzpicture}
	}
    \vspace{-1em}
	\caption{Sensitivity analysis of \ourmodel performance to hyper-parameters. (a) shows the dependency of model performance on temperature $\tau$. (b) illustrates the influence of loss weight $\lambda$. (c) examines the effect of list length $L$.}
    \vspace{-0.8em}
	\label{fig:hyper_para}
\end{figure*}

\subsubsection{Performance w.r.t. Model Size of PLM}
To verify the adaptability and effectiveness of \ourmodel across varying sizes of PLMs, we explore its integration with a range of BERT models, from BERT-mini to BERT-large. This analysis aims to determine whether \ourmodel could generally enhance tool retrieval performance across different model sizes.
Figure~\ref{fig:analy_plmsize} shows that while the performance of the base model naturally improves with larger PLM sizes, the integration of \ourmodel consistently boosts performance across all sizes.
Remarkably, even BERT-mini equipped with \ourmodel, significantly outperforms a much larger BERT-large model (30x larger) operating without our \ourmodel.
These results underscore the generalization and robustness of \ourmodel, demonstrating its potential to significantly improve tool retrieval performance for PLMs of any scale.

\subsubsection{Performance w.r.t. Different Tool Sizes.}
The \ourdata dataset encompasses queries that require $1 \sim 3$ tools, while ToolBench includes queries needing $2 \sim 4$ tools.  To assess how well \ourmodel adapts to queries with diverse tool requirements, we divide each dataset into three subsets based on the number of tools required by each query and conduct a focused analysis on these subsets.
As shown in Figure~\ref{fig:analy_toolsize}, there is a discernible decline in performance as the number of ground-truth tools increases, reflecting the escalating difficulty of achieving complete retrieval. 
However, \ourmodel demonstrates consistent performance improvement across all subsets and backbones. This improvement is especially significant in the most challenging cases, where queries may involve using three or four tools.
These results consistently highlight the robustness of \ourmodel and its potential to meet the complex demands of tool retrieval tasks across various scenarios.

\subsubsection{Hyper-parameter Analysis.}
Figure~\ref{fig:hyper_para} illustrates the sensitivity of \ourmodel to the temperature parameter $\tau$ and the loss weight $\lambda$, but shows relative insensitivity to variations in the sampled list length $L$. The influence of $\tau$ varies across two datasets, suggesting that its impact depends on the specific data distribution. Conversely, the pattern observed for $\lambda$ across both datasets is consistent, marked by an initial performance improvement that eventually plateaus, underscoring the importance of carefully selecting $\lambda$ to maximize the effectiveness of \ourmodel.

\section{Conclusion}
This study introduces \ourmodel, a novel model-agnostic approach designed to enhance the completeness of tool retrieval tasks, comprising two stages: semantic learning and collaborative learning.
We initially employ semantic learning to ensure semantic representation between queries and tools.
Subsequently, by incorporating graph collaborative learning and cross-view contrastive learning, \ourmodel captures the collaborative relationships among tools.
Extensive experimental results and analysis demonstrate the effectiveness of \ourmodel, especially in handling multifaceted queries with multiple tool requirements. 
Furthermore, we release a new dataset \ourdata and introduce a novel evaluation metric \oureval, both of which are valuable resources for facilitating future research on tool retrieval.

\begin{acks}
This work was funded by the National Key R\&D Program of China (2023YFA1008704), the National Natural Science Foundation of China (No. 62377044), Beijing Key Laboratory of Big Data Management and Analysis Methods, Major Innovation \& Planning Interdisciplinary Platform for the ``Double-First Class'' Initiative, funds for building world-class universities (disciplines) of Renmin University of China, and PCC@RUC.
\end{acks}

\clearpage
\bibliographystyle{ACM-Reference-Format}
\balance
\bibliography{ref}

\end{sloppy}
\end{document}